\pdfoutput=1
\documentclass{article}

\usepackage{arxiv}
\usepackage[utf8]{inputenc}
\usepackage[T1]{fontenc}
\usepackage{microtype}
\usepackage{amsmath,amssymb}
\usepackage{booktabs}
\usepackage{multirow}
\usepackage{makecell}
\usepackage{threeparttable}
\usepackage{array}
\usepackage{tabularx}
\usepackage{graphicx}
\usepackage{xcolor}
\usepackage[numbers,sort&compress]{natbib}
\usepackage{url}
\usepackage[hidelinks]{hyperref}
\usepackage[capitalise,noabbrev]{cleveref}
\usepackage{caption}
\usepackage{enumitem}
\usepackage{listings}
\usepackage{placeins}
\usepackage{wrapfig}

\graphicspath{{figures/}}
\renewcommand{\topfraction}{0.68}
\renewcommand{\bottomfraction}{0.68}
\renewcommand{\textfraction}{0.32}
\renewcommand{\floatpagefraction}{0.9}
\setlist{nosep,leftmargin=*}

\newcommand{\figorbox}[3]{%
  \IfFileExists{figures/#1.pdf}{\includegraphics[width=#2]{#1.pdf}}{%
    \IfFileExists{figures/#1.png}{\includegraphics[width=#2]{#1.png}}{%
      \fbox{\parbox[c][#3][c]{\dimexpr#2-2\fboxsep-2\fboxrule\relax}{\centering\small
        \textcolor{red}{Figure \texttt{\detokenize{#1}} not yet available (see \texttt{FIGURES.md}).}}}}}}

\newcommand{\figorboxh}[4]{%
  \IfFileExists{figures/#1.pdf}{\includegraphics[width=#2,height=#3,keepaspectratio]{#1.pdf}}{%
    \IfFileExists{figures/#1.png}{\includegraphics[width=#2,height=#3,keepaspectratio]{#1.png}}{%
      \fbox{\parbox[c][#4][c]{\dimexpr#2-2\fboxsep-2\fboxrule\relax}{\centering\small
        \textcolor{red}{Figure \texttt{\detokenize{#1}} not yet available (see \texttt{FIGURES.md}).}}}}}}

\newcommand{\tabinput}[1]{%
  \IfFileExists{tables/#1.tex}{\begingroup
    \renewenvironment{table}[1][]{\centering}{}\renewenvironment{table*}[1][]{\centering}{}%
    \renewcommand{\caption}[2][]{}\renewcommand{\label}[1]{}%
    \input{tables/#1}\endgroup}{%
    \fbox{\parbox[c][2.5cm][c]{0.9\linewidth}{\centering\small
      \textcolor{red}{Table \texttt{\detokenize{#1}} not yet available (see \texttt{FIGURES.md}).}}}}}

\newcommand{\taurus}{Taurus}
\newcommand{\ffs}{Free-form Sonnet}
\newcommand{\ffo}{Free-form Opus}
\newcommand{\autoA}{Taurus-base (auto)}
\newcommand{\revA}{Taurus-base}
\newcommand{\autoB}{Taurus-tuned (auto)}
\newcommand{\revB}{Taurus-tuned}
\newcommand{\autoC}{Taurus-hybrid (auto)}
\newcommand{\revC}{Taurus-hybrid}
\newcommand{\bel}{B11 alone}
\newcommand{\tsg}{TotalSegmentator}
\newcommand{\nbu}{Reference-seeded (not blind)}

\newif\iffullappendix
\ifdefined\FullAppendix\fullappendixtrue\fi

\newif\ifshowsrc
\showsrcfalse
\DeclareRobustCommand{\src}[1]{\ifshowsrc Source: #1.\fi}

\title{Taurus: LLM-in-the-Loop CT Annotation Matches\\Coding Agents at Half the Cost}

\author{Bing Hu \qquad Rishabh Sharma \\[0.45em]
  \normalfont\normalsize David R. Cheriton School of Computer Science, University of Waterloo \\
  \normalfont\normalsize Waterloo, Ontario, Canada}

\renewcommand{\shorttitle}{Taurus: LLM-in-the-Loop CT Annotation}
\renewcommand{\headeright}{Preprint}
\renewcommand{\undertitle}{}
\date{}
\makeatletter
\renewcommand{\@maketitle}{%
  \vbox{%
    \hsize\textwidth
    \linewidth\hsize
    \vskip 0.1in
    \@toptitlebar
    \centering
    {\LARGE\sc \@title\par}
    \@bottomtitlebar
    \begin{tabular}[t]{c}\bf\rule{\z@}{24\p@}\@author\end{tabular}%
    \vskip 0.25in
  }
}
\makeatother

\hypersetup{pdftitle={Taurus: LLM-in-the-Loop CT Annotation Matches Coding Agents at Half the Cost},
            pdfauthor={Bing Hu, Rishabh Sharma}}

\begin{document}
\maketitle

\begin{abstract}
Dense labels of the aorta in CT are slow to draw, and the public aortic CT datasets that could stand in for them each label different structures under their own conventions.
Large language model (LLM) agents can now write code and read images.
We asked whether such a model labels these scans better as a free-form agent that writes its own segmentation code, or inside a harness, where an automated pass labels the scan and the model reviews the result.
We built a harness, \taurus{}, and compared it with free-form agents by blind label recovery: we withheld the human labels of six public datasets (939 volumes) and scored every approach against each dataset's own classes. %
In the first version of \taurus{}, the model's review raised the primary metric on only two of six datasets, by less than the noise between two reviews of the same case. %
A free-form agent running Claude Opus 5.5 scored higher on most datasets. %
Moving what the free-form agents did well into the automated pass then carried most of the gain on 38 held-out cases. %
The final version smooths the label surfaces with a fixed rule and checks for the large errors the automated pass tends to make, such as an aneurysm drawn to half its size.
Each check that fires puts a multiple-choice question to the model, with candidate corrections drawn on the scan. %
It matched the surface smoothness of the human references, and in the primary metric we found no difference from the free-form Opus agent, at under half its cost, \$1.22 against \$2.71 per case. %
Most of what made \taurus{} competitive came from the harness and its automation rather than from the language model itself.
\taurus{} shows that this kind of annotation can be automated, so that expert time can move from drawing labels to checking them.
\end{abstract}


\section{Introduction}
\label{sec:intro}

\begin{wrapfigure}[18]{r}{0.5\textwidth} %
  \centering
  \vspace{-1.2\baselineskip}
  \figorbox{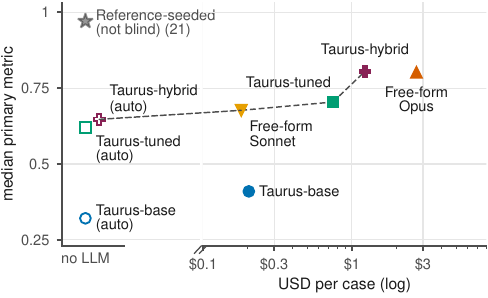}{\linewidth}{5.0cm}
  \caption{\revC{} ends the cost--accuracy front on the 38 held-out cases: median primary metric, pooled for description only, against model cost per case.} %
  \label{fig:pareto}
  \vspace{-0.8\baselineskip}
\end{wrapfigure}

Dense labels of the aorta and its branches are the input to aneurysm sizing, endograft planning and the training of segmentation models, but they are slow to draw.
Public datasets help less than their number suggests, because each one labels a different subset under its own conventions, from 23 named vessels and zones \citep{imran2025aortaseg24} to a single binary lumen tree \citep{radl2022avt} (\cref{tab:datasets}) \citep{siriapisith2022peerj,mosmeddata_aaa,jamshidian2025fourdcta,wilson2013vmr}. %
Combining them means relabeling them under one label contract, which in this study has four tissue classes (lumen, calcium, thrombus and wall) and 27 vessel labels for the ten aortic zones and the named branches (\cref{sec:protocol}).

\begin{figure}[t]
  \centering
  \figorboxh{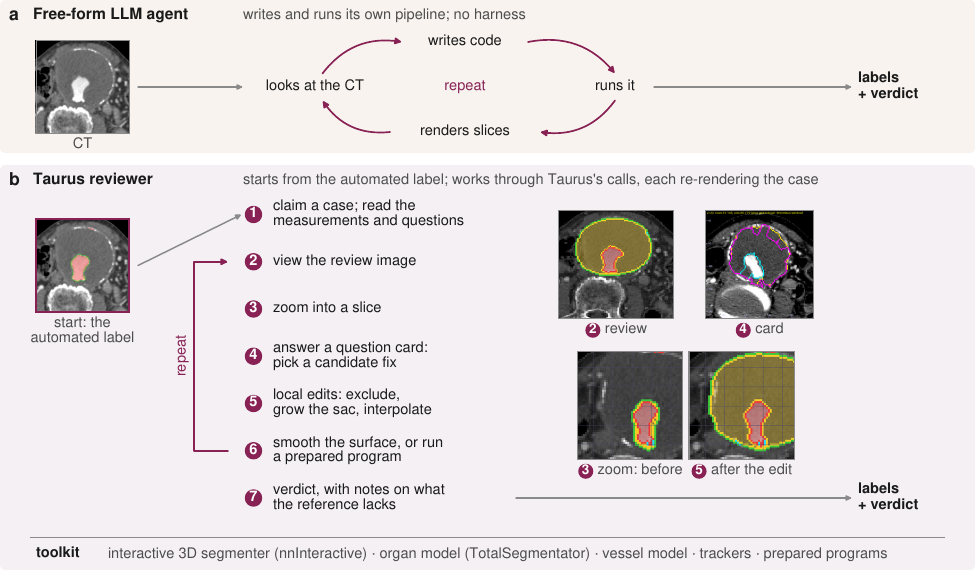}{0.95\linewidth}{\textheight}{5.6cm}\\[1pt]
  {\setlength{\unitlength}{1in}\begin{picture}(6.175,1.28)
    \put(0,0){\includegraphics[width=0.95\linewidth,trim=0 24bp 0 0,clip]{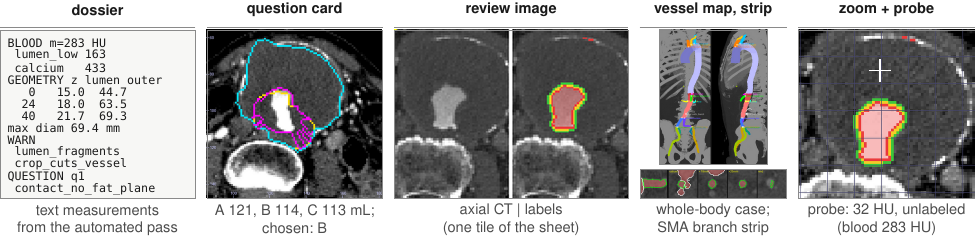}}
    \put(0.02,1.19){\sffamily\fontsize{8.5}{10}\selectfont\textbf{c}}
  \end{picture}}
  \caption{Two ways to let a language model label a CT scan.
  (a) The free-form agent loops through looking, writing code, running it and rendering until it writes the labels and a verdict.
  (b) The \taurus{} reviewer (\revC{} shown) starts from the automated label and works through harness calls, each re-rendering the case, until it records a verdict.
  (c) What the reviewer sees; the question card offers three outlines of the aneurysm sac (A 121, B 114, C 113\,mL), and the reviewer chose B.}
  \label{fig:overview}
\end{figure}

Supervised and promptable segmentation models \citep{isensee2021nnunet,wasserthal2023totalseg,isensee2025nninteractive} produce good masks for the classes they know, but someone still has to check the output and decide what to fix.
Large language model (LLM) agents can now run code, call tools and read images \citep{yao2023react,yang2024sweagent}, so a model could take over both the labeling and its checking, in one of two ways (\cref{fig:overview}).
A \emph{free-form} agent receives the scan and a programming environment and writes its own segmentation code.
A \emph{harness}, here our system \taurus{}, runs an automated pass and then gives the model renders and edit commands, and this model, the \taurus{} reviewer, decides what is wrong, how to fix it and whether the case needs a human.
When the automated pass meets a decision it cannot settle, or one of its checks finds a likely error, the harness puts a multiple-choice question to the reviewer, which we call a \emph{question card}.
For example, a card may show one slice of an aneurysm with three candidate outlines of the sac (121, 114 and 113\,mL), and the reviewer answers with a letter that the harness then applies, so the model chooses among prepared fixes instead of drawing them (\cref{fig:overview}c). %
We ask which of the two labels better and at what cost, and whether the model's verdict tells a human where to look.
Because we had no new expert labels, we used \emph{blind label recovery}: each approach sees only the CT and is scored against the hidden human labels of six public datasets. %
These datasets label different structures, so before scoring we convert every output to the classes of the dataset at hand, which we call its label space.

\begin{table}[t]
  \centering
  \caption{The six datasets label different structures, from 23 vessel ids to a single envelope mask, and each is scored against its own classes.
  The reference image shows the human label on one design-set case, and contrast is given over the reviewed cases.}
  \label{tab:datasets}
  \scriptsize\let\footnotesize\scriptsize %
  \begin{table}[t]
\centering
\footnotesize
\caption{Datasets. Every dataset is scored in its own label space against its own human label, which the approaches never see.}
\label{tab:datasets}
\setlength{\tabcolsep}{3pt}
\begin{tabular}{@{}>{\centering\arraybackslash}m{1.1cm} >{\raggedright\arraybackslash}m{2.0cm} >{\raggedright\arraybackslash}m{3.2cm} >{\raggedright\arraybackslash}m{2.5cm} >{\raggedright\arraybackslash}m{1.15cm} >{\centering\arraybackslash}m{2.55cm} >{\raggedright\arraybackslash}m{2.1cm}@{}}
\toprule
Reference & Dataset & Human label (type) & Field of view & Contrast & \makecell{Volumes / reviewed /\\[-1pt]free-form / held-out} & Primary metric \\
\midrule
\includegraphics[height=0.95cm]{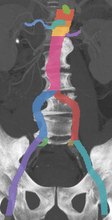} & AortaSeg24 \citep{imran2025aortaseg24} & 23 vessel ids (zones 0--9, branches) (voxel label map) & whole body; type-B dissections & CTA & 100 / 100 / 5 / 4 & abdominal vessel macro Dice \\ \addlinespace[2pt]
\includegraphics[height=0.95cm]{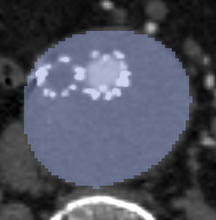} & Siriapisith AAA/EVAR \citep{siriapisith2022peerj} & envelope\textsuperscript{a} (voxel label map) & 80 mm abdominal crops, 256 $\times$ 256 $\times$ 64; incl. EVAR & \makecell[l]{CTA;\\[-1pt]10 NC} & 520 / 160 / 6 / 12 & outer Dice \\ \addlinespace[2pt]
\includegraphics[height=0.95cm]{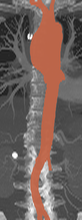} & AVT \citep{radl2022avt} & one binary arterial-tree lumen (voxel segmentation) & thorax--pelvis arterial tree & \makecell[l]{CTA;\\[-1pt]1 NC} & 55 / 24 / 5 / 4 & tree Dice \\ \addlinespace[2pt]
\includegraphics[height=0.95cm]{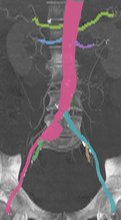} & VMR \citep{wilson2013vmr} & lumen per named vessel (lofted surface models, voxelized) & abdomen ($\pm$ chest); 16 AAA, 1 healthy & CTA & 17 / 17 / 4 / 4 & per-vessel macro Dice \\ \addlinespace[2pt]
\includegraphics[height=0.95cm]{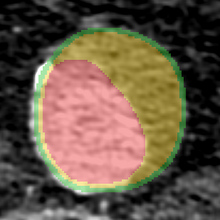} & Jamshidian 4D-CTA \citep{jamshidian2025fourdcta} & thrombus, sac and wall meshes (finite-element surface meshes, voxelized) & aneurysm-sac crops & CTA & 149\textsuperscript{b} / 39 / 5 / 6 & thrombus Dice \\ \addlinespace[2pt]
\makecell{\textit{not shown:}\\[-1pt]ND licence\textsuperscript{c}} & MosMedData AAA \citep{mosmeddata_aaa,kodenko2025dataset} & lumen, calcium (thrombus on 4) (voxel label map) & abdominal aortic segment (104--209 mm) & CTA & 98 / 98 / 5 / 8 & lumen + calcium mean Dice \\ \addlinespace[2pt]
\midrule
& Total & & & & 939 / 438 / 30 / 38 & \\
\bottomrule
\end{tabular}
\par\smallskip\footnotesize\raggedright \textsuperscript{a}the envelope is one mask that covers the whole aneurysm, lumen, thrombus and wall, out to its outer surface. \textsuperscript{b}cardiac phases of 20 patients, scored per phase. \textsuperscript{c}the no-derivatives (ND) term of its CC BY-NC-ND 3.0 licence does not allow transformed images. Reviewed = cases with a Taurus-base verdict (main experiment); free-form = main cases run by the free-form agents; held-out = the evaluation cases of the held-out comparison. NC = non-contrast among the reviewed cases. VMR image data \copyright{} Stanford University, the Regents of the University of California, Open Source Medical Software Corporation, and other parties.\par
\end{table}

\end{table}

We built three versions of the harness and compared each with free-form agents running the same models.
\revA{} uses fixed thresholds and lets the reviewer only correct what the pass drew, and \revB{} moves the steps with which the free-form agents won into the automated pass.
\revC{} adds a surface regularization, a fixed step that smooths the label surfaces and places the lumen edge at a set fraction of the blood's brightness.
It also lets the reviewer refine edges with short programs, and it checks the scored classes for gross errors, each of which it turns into a question card.
Most of what made \taurus{} competitive turned out to be deterministic: the free-form agent's winning steps, once automated, carried most of the held-out gain, and a deterministic regularization made the boundaries as smooth as the references. %
Given these cards, the model raised the held-out median from 0.65 to 0.81 and left no detectable difference from the free-form agent, at under half its cost (\cref{fig:pareto}). %
Its verdict, however, was a weak signal of failure, and models trained on labels from \revA{} fell well short of models trained on human labels. %

\section{Related work}
\label{sec:related}

Self-configuring networks such as nnU-Net remain strong baselines for 3D medical segmentation \citep{isensee2021nnunet,isensee2024revisited}.
TotalSegmentator segments over 100 CT structures including the aorta \citep{wasserthal2023totalseg}, and vesselFM targets vessels across modalities \citep{wittmann2025vesselfm}. %
Promptable models extend Segment Anything \citep{kirillov2023sam,ravi2024sam2} to medical images in 2D \citep{ma2024medsam,wong2024scribbleprompt} and 3D \citep{wang2025sammed3d,du2024segvol,he2025vista3d,isensee2025nninteractive}, and MONAI Label puts such models into an annotation tool \citep{diazpinto2024monailabel}.
All of them assume a person who places the prompts and judges the result, and \taurus{} asks a language model to take that person's place.
For the aorta itself, networks segment the aneurysm lumen and intraluminal thrombus (ILT) on CT angiography (CTA) \citep{lopezlinares2018thrombus,lareyre2019pipeline,caradu2021praevaorta,derycke2025unet}, and challenges have moved to the whole vessel tree \citep{jin2025sega,imran2025aortaseg24}, with zones that follow the Society for Vascular Surgery (SVS) reporting standards \citep{fillinger2010tevar,lombardi2020tbad}.

SWE-agent showed that the interface an agent works through changes how well a fixed model solves a task \citep{yang2024sweagent}, and a harness is such an interface for annotation.
In medicine, vision-language models answer questions about images \citep{li2023llavamed,tu2024medpalmm}, agents call imaging tools \citep{li2024mmedagent,fallahpour2025medrax,hoopes2024voxelprompt}, and language models serve as annotators and judges \citep{gilardi2023chatgpt,zheng2023judging}.
Recent agents also drive promptable segmenters directly \citep{liu2025medsam3,liu2026medsamagent}, whereas we study a model that reviews and edits the output of a deterministic labeler and compare it with an agent that writes its own code.
The closest evaluation asks multimodal models whether 2D segmentation overlays are acceptable \citep{sheng2025beyondpixel}, and we know of no study that gives an agent dense 3D CT labeling end to end and measures it against independent human labels together with its cost.
Segmentation quality can also be predicted without ground truth \citep{kohlberger2012errors,valindria2017rca,robinson2019qc}, and we test the reviewer's verdict as such a prediction.

\section{Method}
\label{sec:method}

\taurus{} runs an automated pass, which proposes the vessels, grows the trunk and sac, assigns the tissue classes, names the branches and zones and raises quality flags, and then renders its labels for the \taurus{} reviewer. %
This reviewer is an LLM session that edits the labels only through a fixed set of edit commands and closes each case with one of three verdicts: \emph{ok}, \emph{fixed} or \emph{issue}, the last meaning that the case needs a human (\cref{fig:overview}b). %
Reviewers read the harness guide (\cref{app:prompts}) and see only the images that harness commands return. %

\begin{table}[t]
  \centering
  \caption{The harness toolkit and the \taurus{} versions that have each tool ($\checkmark$: yes; ($\checkmark$): in limited form; --: no).}
  \label{tab:tools}
  \scriptsize
  \begin{tabularx}{\linewidth}{@{}l X c c c@{}}
    \toprule
    Tool & What it does & \revA{} & \revB{} & \revC{} \\
    \midrule
    \multicolumn{5}{@{}l}{\textit{Automated pass}} \\
    Proposal & AI segmentation models (vessel model, whole-body organ model) or a bright core propose the aorta and branches & $\checkmark$ & $\checkmark$ & $\checkmark$ \\
    Trunk and sac growth & grows the trunk and sac; \revB{} and \revC{} add a trunk tracker and a contour out to the aneurysm's outer surface, where its wall meets the surrounding fat; \revC{} re-seeds the trunk when the labeled lumen is implausibly small or a brighter tube is left unlabeled, and checks the sac against a polar edge finder & $\checkmark$ & $\checkmark$ & $\checkmark$ \\
    Tissue classes & lumen, calcium, thrombus and wall from thresholds, which are fixed in \revA{} and measured on the case in \revB{} and \revC{}; \revC{} drops calcium that lies on the vertebral cortex & $\checkmark$ & $\checkmark$ & $\checkmark$ \\
    Branches, zones, flags & traces branches from the wall, names them and the aortic zones, raises quality warnings; \revC{} adds warnings for a rebuilt trunk, an under-sized sac, calcium specks and a header orientation that contradicts the anatomy & $\checkmark$ & $\checkmark$ & $\checkmark$ \\
    \midrule
    \multicolumn{5}{@{}l}{\textit{What the reviewer sees}} \\
    Renders & review image, vessel map, branch strips, zoomed slices; attenuation probes from \revB{} on & $\checkmark$ & $\checkmark$ & $\checkmark$ \\
    Dossier and cards & measurements as text; decisions posed as candidate masks to choose from; \revC{} adds cards for the sac extent, calcium specks, a rebuilt trunk and the orientation, each stating its stake in the dataset's scored classes & -- & $\checkmark$ & $\checkmark$ \\
    \midrule
    \multicolumn{5}{@{}l}{\textit{Edit commands}} \\
    Paint, erase, clear & local corrections of any class, removal of detached pieces & $\checkmark$ & $\checkmark$ & $\checkmark$ \\
    AI outer growth & click-seeded growth of the outer boundary with the interactive 3D segmenter & $\checkmark$ & $\checkmark$ & $\checkmark$ \\
    Lumen threshold & case-wide lower lumen threshold and lumen growth from a seed & $\checkmark$ & $\checkmark$ & $\checkmark$ \\
    Sac growth, exclusion & grows the sac out to the surrounding fat, interpolates between slices, excludes drawn regions & -- & $\checkmark$ & $\checkmark$ \\
    Trunk rebuild & re-tracks the trunk from a seed & -- & $\checkmark$ & $\checkmark$ \\
    Branch edits & add, extend, remove and rename vessels & $\checkmark$ & $\checkmark$ & $\checkmark$ \\
    Regularization & when the case is claimed, moves the lumen edge to 0.6 times blood attenuation, smooths the outer surface and redraws the wall as a thin ring & -- & -- & $\checkmark$ \\
    Prepared and own programs & prepared programs, written and tested in advance, such as mask smoothing, edge refinement, branch tracing or the polar sac finder (19 in \revC{}), or the reviewer's own short programs & -- & ($\checkmark$) & $\checkmark$ \\
    Undo and verdict & undo any edit; \emph{ok}, \emph{fixed} or \emph{issue} with codes; reference-disagreement notes in \revC{} & $\checkmark$ & $\checkmark$ & $\checkmark$ \\
    \bottomrule
  \end{tabularx}
\end{table}

\begin{figure}[t]
  \centering
  \figorboxh{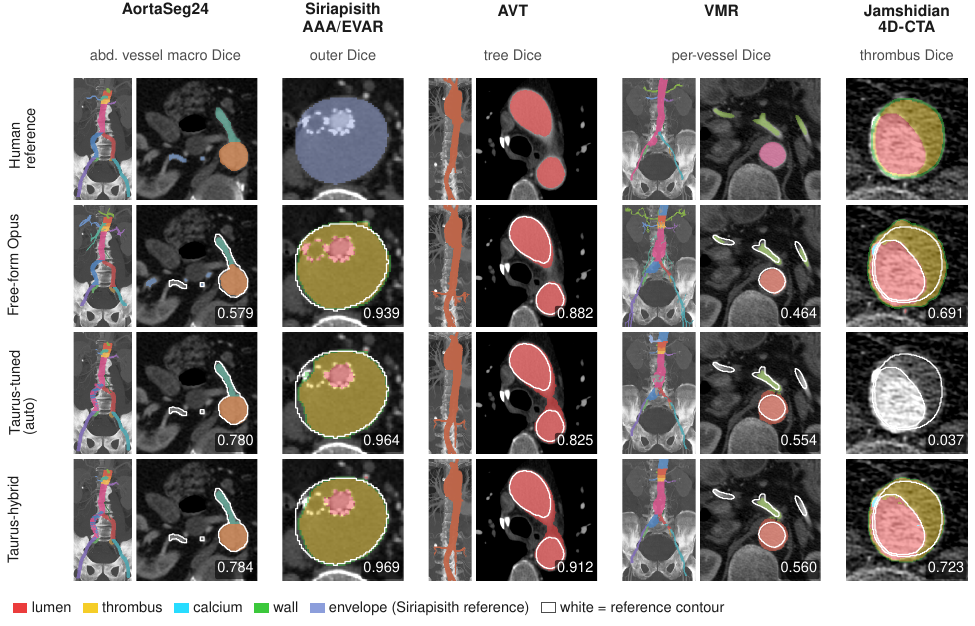}{\linewidth}{\textheight}{5.0cm} %
  \caption{Example labels from the design set, one case per dataset among the 30 free-form cases.
  Rows show the human reference, \ffo{}, \autoB{} (\revB{} was not reviewed on these cases) and \revC{} on the CT, with the reference contour as a white line, vessel ids in segment colors and each case's primary metric in the corner.
  AortaSeg24, AVT and VMR are shown as a coronal projection with an axial slice through the trunk and its branches.
  The Siriapisith AAA/EVAR reference is an envelope, one mask that covers the whole aneurysm, lumen, thrombus and wall, out to its outer surface.
  For AortaSeg24 and Jamshidian 4D-CTA we show the case with the median primary metric of the first hybrid build, and for Siriapisith AAA/EVAR, AVT and VMR the case with its largest margin over \ffo{}.
  The \ffo{} label volumes of the other design-set cases were not kept, so the cases of the first build are shown.
  MosMedData AAA is not shown, because its licence (CC BY-NC-ND 3.0) does not allow transformed images.
  VMR data \copyright{} Stanford University, the Regents of the University of California, Open Source Medical Software Corporation, and other parties.} %
  \label{fig:overlays}
\end{figure}

\begin{table}[t]
  \centering
  \caption{On the held-out cases, \revC{} was ahead of both free-form agents on five of six datasets, and \ffs{} led on AVT, where the automated pass found no usable trunk.
  Median primary metric per dataset (each in its own metric, so the pooled column is descriptive) and over all 38 cases, and model cost and minutes per case; for \taurus{} the minutes include the vessel proposal and the automated pass (\cref{app:compute}).
  Bold marks the highest median of the column, and asterisks mark medians that rest on two near-empty labels and on the calcium scoring convention (\cref{sec:res-v2}).
  Spreads, the hard and typical cases and turns are in \cref{tab:a5}.} %
  \label{tab:heldout}
  \footnotesize
  \begin{table*}[t]
\centering
\footnotesize
\caption{Held-out evaluation set (38 cases): median primary metric per dataset and over all cases, and model cost and minutes per case from scan to label (automated pass: vessel proposal on the GPU and the pass on CPU; review: that plus the review). Bold = the highest median of the column. Each dataset keeps its own primary metric (Table~\ref{tab:datasets}).}
\label{tab:heldout}
\setlength{\tabcolsep}{3pt}
\begin{tabular*}{\textwidth}{@{\extracolsep{\fill}}lccccccccc@{}}
\toprule
 & \multicolumn{6}{c}{Primary metric per dataset} & All & \multicolumn{2}{c}{Per case} \\
\cmidrule(lr){2-7}\cmidrule(l){9-10}
Approach & \makecell{AortaSeg24\\[-1pt](4)} & \makecell{Siriapisith\\AAA/EVAR\\[-1pt](12)} & \makecell{MosMedData\\AAA\\[-1pt](8)} & \makecell{AVT\\[-1pt](4)} & \makecell{VMR\\[-1pt](4)} & \makecell{Jamshidian\\4D-CTA\\[-1pt](6)} & \makecell{all\\[-1pt](38)} & \$ & min \\
\midrule
Taurus-base (auto) & 0.572 & 0.496 & 0.021 & 0.027 & 0.490 & 0.002 & 0.321 & -- & 0.7 \\
Taurus-base & 0.579 & 0.496 & 0.024 & 0.432 & \textbf{0.492} & 0.004 & 0.410 & 0.20 & 1.5 \\
Taurus-tuned (auto) & 0.732 & 0.783 & 0.462 & 0.437 & 0.464 & 0.534 & 0.620 & -- & 1.0 \\
Taurus-tuned & 0.732 & 0.871 & 0.462 & 0.437 & 0.466 & 0.572 & 0.704 & 0.75 & 6.4 \\
Taurus-hybrid (auto) & \textbf{0.734} & 0.791 & 0.468 & 0.454 & 0.468 & 0.587 & 0.647 & -- & 1.1 \\
Taurus-hybrid & 0.733 & \textbf{0.897} & \textbf{0.853} & 0.442 & 0.483 & \textbf{0.732} & \textbf{0.805} & 1.22 & 6.4 \\
\midrule
Free-form Sonnet & 0.307 & 0.802 & 0.773$^{*}$ & \textbf{0.823}$^{*}$ & 0.226 & 0.430 & 0.677 & 0.18 & 1.3 \\
Free-form Opus & 0.629 & 0.892 & 0.837 & 0.466$^{*}$ & 0.446 & 0.715 & 0.802 & 2.71 & 16.1 \\
\bottomrule
\end{tabular*}
\end{table*}
\end{table}

\begin{figure}[t]
  \centering
  \figorboxh{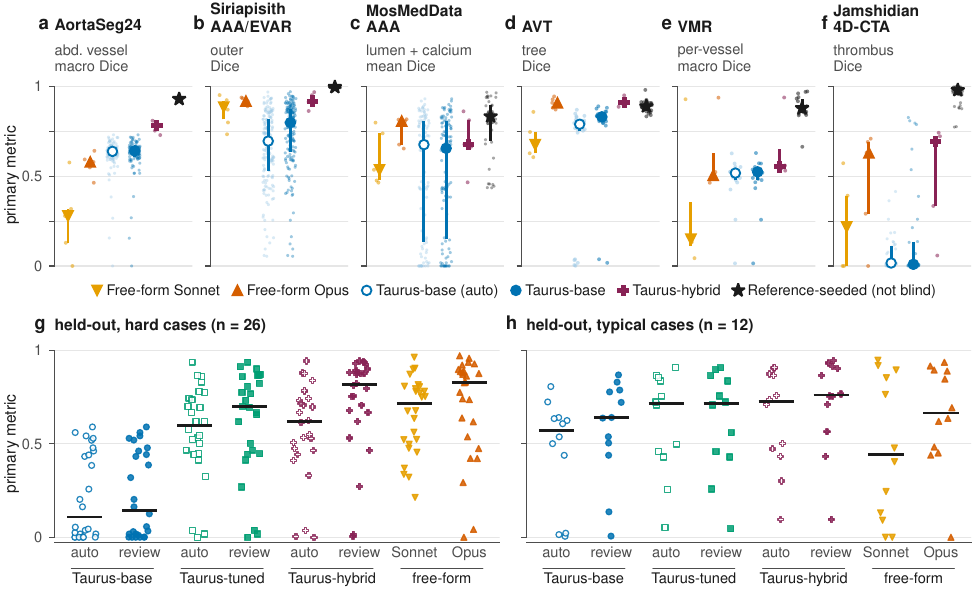}{\linewidth}{\textheight}{3.6cm}
  \caption{Review raised \revA{} on two datasets by small amounts; on the held-out cases \autoB{} carried the gain of \revB{}, and the cards that of \revC{}.
  Primary metric per approach.
  (a--f) Main experiment per dataset (marker: median, bar: interquartile range, dots: cases), including \revC{} on the 30 free-form cases, which are design-set cases for it.
  Each panel uses its dataset's own metric and label space, so values are not comparable across panels.
  (g, h) Held-out hard and typical cases, each in its own dataset's metric and pooled for description only.
  Horizontal bars and numbers are medians.
  }
  \label{fig:main}
\end{figure}

\begin{figure}[t]
  \centering
  \figorbox{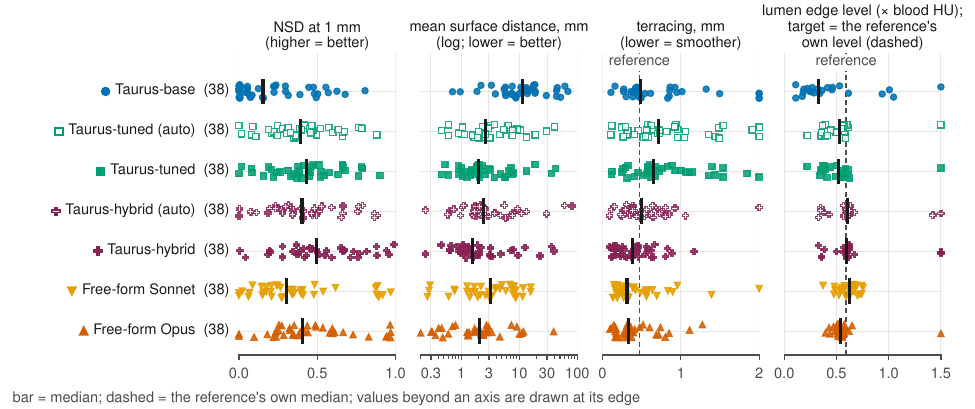}{\linewidth}{4.5cm}
  \caption{The regularization smoothed the surfaces.
  Boundary and surface quality against the human reference on the 38 held-out cases: NSD at 1\,mm, MSD (log scale), terracing index and lumen edge level per approach.
  Bars are medians, dots are cases, and points beyond an axis are drawn at its edge.
  The dashed lines mark the reference's own median terracing and lumen edge level.}
  \label{fig:surface}
\end{figure}

\begin{figure}[t]
  \centering
  \figorboxh{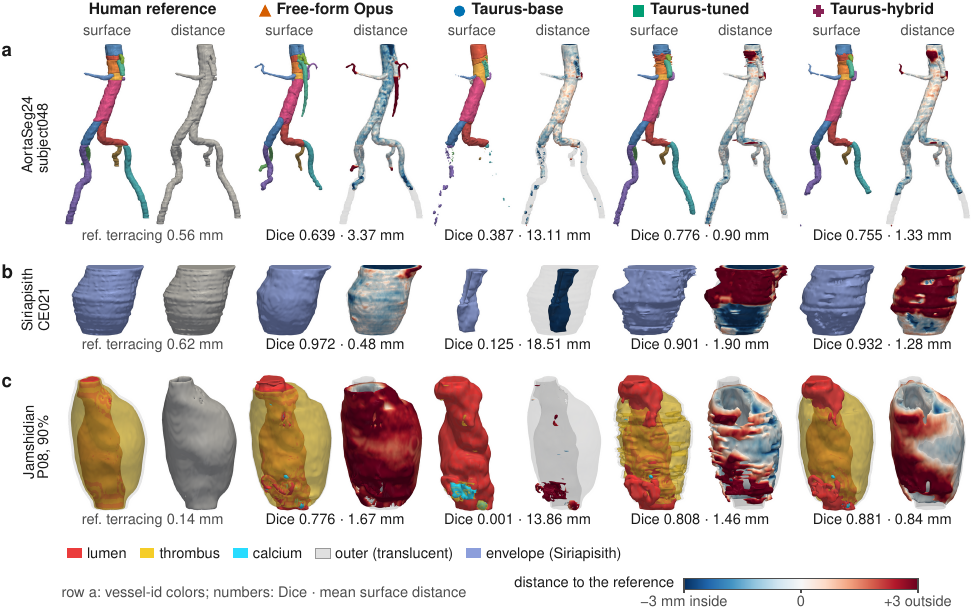}{\linewidth}{\textheight}{12cm}
  \caption{\revA{} misses the sac or thrombus, \revB{} restores them with slab terraces, and \revC{} removes the terraces.
  Mesh-to-mesh comparison with the human reference on three held-out cases: (a) the AortaSeg24 vessel tree, (b) a Siriapisith AAA/EVAR outer envelope, (c) a Jamshidian 4D-CTA thrombus.
  The upper row shows each approach's surface.
  The row beneath colors it by signed distance to the reference, blue inside and red outside on a fixed $\pm3$\,mm scale, with the reference translucent.
  The Human column prints the reference's own terracing index, and the others print the primary metric and the MSD.}
  \label{fig:meshes}
\end{figure}

\begin{figure}[t]
  \centering
  \figorboxh{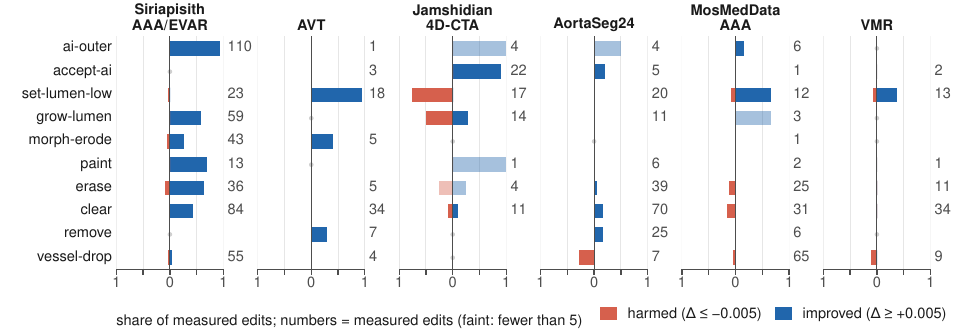}{\linewidth}{\textheight}{4.0cm}
  \caption{Most edits of the \revA{} reviewers left the primary metric unchanged, and two commands carried the gain.
  Replay of the \revA{} edits in the main experiment: for each edit command and dataset, the share of measured edits that raised (blue) or lowered (red) the case's primary metric by at least 0.005.
  Numbers are measured edits, and faint bars have fewer than five.
  \texttt{ai-outer} grows the outer boundary with the interactive segmenter, and \texttt{set-lumen-low} lowers the lumen threshold.}
  \label{fig:edits}
\end{figure}

\subsection{\taurus{}: LLM-in-the-Loop Annotations}
\label{sec:protocol}

\taurus{} comes in three versions, \revA{}, \revB{} and \revC{}, which share the automated-pass stages, the renders, the verdict and the segmentation models.
A version's name alone denotes its reviewed output, and ``(auto)'' marks the automated pass alone, as in \autoA{}.
In \revA{}, the automated pass uses fixed thresholds, and the reviewer can only correct what the pass drew, by painting or erasing a region, growing the outer boundary from a click, editing branches or lowering the lumen threshold, but cannot re-run the sac growth or the trunk tracking.
We built \revB{} by moving what the free-form agents did well into the automated pass: it measures blood attenuation for its thresholds, draws the aneurysm contour out to its outer surface, tracks the trunk from a seed, and poses decisions as question cards with candidate masks. %

\revB{} labeled accurately but drew stair-stepped surfaces, and its reviewer wrote code of its own only 4 times in 38 cases. %
We built \revC{} in two steps, the first of which kept the automated pass of \revB{} and changed the review to close this surface gap at low cost. %
When the reviewer claims a case, a deterministic regularization moves the lumen edge to 0.6 times the measured blood attenuation, smooths the outer surface and redraws the wall as a thin ring. %
We chose the 0.6 level on references outside the held-out set only. %
In a second review phase, the reviewer inspects the edges on zoomed slices and may refine them with prepared programs, most of which were distilled from code that \ffo{} wrote, or with short programs of its own (\cref{tab:tools}). %
After every edit the reviewer compares the same zoomed slice before and after and undoes the edit if the edge did not move closer to the CT boundary, while the harness itself refuses only edits that remove too much. %

The first build matched the references' smoothness but trailed \ffo{} on the held-out cases. %
The gap sat in a few gross errors in the scored classes, a trunk placed on the spine, sacs drawn to half their extent and calcium specks on calcium-free scans (\cref{sec:res-v2}). %
The second build, the \revC{} of this paper, adds automated checks for these errors and turns each into a question card with a candidate fix. %
All three checks were ported from what \ffo{} had done in its transcripts. %
Each card states its stake in the dataset's scored classes, and the guide tells the reviewer to check those classes at the gross level before any edge work. %
The budget was raised to about 60 turns, 32 image opens and 28 edits to match what \ffo{} spent per case. %
\Cref{tab:tools} lists the toolkit of each version, and \cref{fig:overview}c shows what the reviewer is given to see.

\subsection{Blind Label Recovery}

To keep every approach blind, we built for each dataset a copy that holds only the CT volumes, while the human labels stay with the scorer. %
The starting vessels come from our nnU-Net \citep{isensee2021nnunet} vessel model, applied out-of-fold on AortaSeg24 where it was trained, or from TotalSegmentator \citep{wasserthal2023totalseg} (\cref{app:data}). %
Every approach writes two maps under one label contract: a tissue map with lumen, calcium, thrombus and a generated wall shell, and a vessel map with 27 labels, the ten aortic zones and the named branches (\cref{app:data}). %

\paragraph{Datasets.}
We used six public datasets, AortaSeg24, the Aortic Vessel Tree (AVT), Siriapisith AAA/EVAR, MosMedData AAA, Jamshidian 4D-CTA and the Vascular Model Repository (VMR), whose sources and full names are given in \cref{tab:datasets} and \cref{app:data}. %
Their human labels cover different structures, from 23 named vessels and zones in AortaSeg24 to a single \emph{envelope} in Siriapisith AAA/EVAR, one mask that covers the whole aneurysm, lumen, thrombus and wall, out to its outer surface.

\paragraph{Held-out design.}
\label{sec:v2}
\label{sec:v3}
We tested \revB{} on 38 held-out cases that its development never touched, fixed before we read any transcript: 26 \emph{hard} cases, on which \revA{} scored in the bottom quartile of its primary metric on that dataset, and 12 \emph{typical} cases from its 40th--60th percentile. %
Development used the 30 free-form cases, 8 learning cases and one in-house scan (\cref{app:data}), and the sessions that extracted what worked from the transcripts saw no held-out case. %
Because the hard cases were chosen for a low \revA{} score, we re-ran and re-reviewed \revA{} on them as a paired control.
\revB{} and the first hybrid build start from the same frozen \autoB{} labels. %
\ffs{} and \ffo{} ran on all 38 cases. %
\revC{}, however, is not a blind test: we designed its first build after seeing the held-out results of \revB{}, and its second build after reading the held-out transcripts of the first build and of \ffo{} to find where they differed. %
We tuned the checks and cards on development cases only, and the held-out cases were run once. %
Its held-out results therefore show what such guidance can achieve rather than how a fixed design generalizes.

\subsection{Evaluations}
\label{sec:scoring}
Outputs are converted to each dataset's classes, its label space, and scored with one primary metric per dataset (\cref{tab:datasets}), and the study plan fixed these metrics and the scored region before the first run. %
Four datasets are scored within a margin of the human label, so that structures the reference never covered do not count as errors (\cref{app:data}). %
On MosMedData AAA, the mean of lumen and calcium Dice skips a class absent from both maps, so a few calcium voxels in a calcium-free case halve it. %

\paragraph{Metrics.}
\label{sec:metrics}
The Dice coefficient \citep{dice1945} measures the overlap of two masks, from 0 when they share no voxel to 1 when they are identical, and macro Dice averages it over the classes present in either mask. %
Because Dice is insensitive to boundary errors on large structures, we also report the normalized surface distance (NSD) at 1\,mm \citep{nikolov2021nsd}, the mean surface distance (MSD) and the 95th-percentile Hausdorff distance (HD95) \citep{maierhein2024metrics} over the primary-metric classes. %
The terracing index, the mean deviation of each axial slice's contour from the mean of its two neighbors, measures slab-like steps in a surface. %
The lumen edge level is the attenuation at which a label places the lumen boundary, as a fraction of blood attenuation, and we compute both measures on the reference too. %

To judge the verdict, we call a case failed when its primary metric is below the 25th percentile of the dataset's \revA{} values.
We report the area under the ROC curve (AUROC) \citep{hanley1982auc} of the binary \emph{issue} verdict, of the harness ranking and of the quality flags alone. %
Cohen's $\kappa$ \citep{cohen1960kappa} compares two reviews of the same case, and a replay scores every saved label version, counting an edit as an improvement or a harm when the primary metric moves by at least 0.005. %

\paragraph{Statistics and cost.}
Paired comparisons use the Wilcoxon signed-rank test \citep{wilcoxon1945} with bootstrap 95\,\% confidence intervals (CI) over 1,000 resamples \citep{efron1979bootstrap}. %
Tests in the main experiment stay within a dataset and are Holm-adjusted \citep{holm1979}, while held-out tests pool the 38 cases and are unadjusted, and medians pooled over datasets are descriptive only.
Two reviews of the same case by the same model tier differ with a standard deviation of 0.020, so we treat per-case differences below 0.040 as run-to-run noise and call 0.040 the noise floor. %
Costs are model costs per case in US dollars, computed from the token counts of each run as described in \cref{app:compute}. %

\section{Results}
\label{sec:results}

We report the main experiment and the held-out comparison, with the paired statistics in \cref{tab:pairs}.
In a third experiment, segmentation models trained on \revA{} labels reached a primary metric of 0.699 on Siriapisith AAA/EVAR and 0.522 on MosMedData AAA, against 0.975 and 0.804 when trained on human labels (\cref{app:downstream}). %

\paragraph{Free-form Baseline.}
\label{sec:freeform}
The free-form agent uses the same models without the harness: Claude Sonnet 5.5 for \ffs{} and Claude Opus 5.5 for \ffo{}. %
It works in a sandbox that holds the CT, the label contract and a task description, but no harness code, no trained model and no other case (\cref{app:prompts}). %
Each case gets one scored run, a run without valid output scores zero, and since only \taurus{} has trained models, the comparison is also one of tools. %

In the main experiment, \autoA{} ran on all 939 volumes and \revA{} on 438 sampled cases. %
The free-form agents ran on 30 of these, four to six per dataset, and \revC{} later reviewed the same 30 cases with its final guide, which makes them design-set cases for it (\cref{fig:overlays}). %
\nbu{}, which runs \taurus{} with the dataset's own labels as priors, bounds the loss that comes from mapping between label spaces. %

\subsection{LLM-in-the-Loop Annotation Review}
\label{sec:res-accuracy}
Review raised the median primary metric of \revA{} on two of six datasets by small but consistent amounts, 0.013 in outer Dice on Siriapisith AAA/EVAR and 0.022 in tree Dice on AVT, both below the noise floor for a single case (\cref{fig:main}). %
\ffo{} did better than \revA{} on most cases of four datasets, tied on VMR and did worse on all five AortaSeg24 cases, but with four to six cases per dataset no difference survived Holm correction (\cref{tab:app-pairs}). %
Most other losses of \revA{} came from 17 of 98 MosMedData AAA scans, whose header contradicted the anatomy and put the proposal on the spine, and from under-grown thrombus on the Jamshidian 4D-CTA phases. %
The automated pass either worked or failed outright, which explains the wide ranges in \cref{fig:main}.
MosMedData AAA scans with a flipped header, and Siriapisith AAA/EVAR crops where the pass missed the sac or found no aorta, scored near 0, far below the other cases. %

On the 30 design-set cases, \revC{} did better than \ffo{}, with a paired median lead of 0.027 in the primary metric (21 better, 6 worse; $p=0.001$, unadjusted). %
The lead in NSD at 1\,mm was larger, 0.095 (23/6; $p<0.001$), and it came at \$1.20 per case against \$4.47. %
It led clearly on AortaSeg24, with an abdominal vessel macro Dice of 0.784 against 0.579, and on the Jamshidian 4D-CTA phases (5/0). %
It differed little from \ffo{} on AVT and the Siriapisith AAA/EVAR crops and trailed on MosMedData AAA (2/3), where the gap came from the calcium artifact (\cref{sec:scoring}). %

\subsection{Learning Automation from Free-Form Annotations}
\label{sec:res-v2}
\label{sec:res-v3}
On the held-out cases, \autoB{} exceeded \autoA{} by a paired median of 0.258 (\cref{fig:main}g--h, \cref{tab:heldout}). %
Review on top of \autoB{} changed only 12 of the 38 cases, and a second review of \autoA{} also changed 12.
In \revB{} the gain therefore came from the automated pass, which applies as fixed rules the steps that the free-form agents had worked out for each dataset. %
The first hybrid build changed little more: its review moved the paired median by 0.001 and the pooled median from 0.647 to 0.716 (\cref{tab:pairs}). %

In \revC{}, by contrast, the review carried the gain, through the cards: the pooled median rose from 0.647 for \autoC{} to 0.805, with 24 cases improved and 5 worse (paired median $+0.025$; $p<0.001$; \cref{tab:heldout}). %
The reviewers took the sac finder's contour on 9 cases, cleared the calcium specks on 5 and rebuilt the trunk out of the vertebral canal on the header-flipped MosMedData AAA scan. %
The largest gains came on MosMedData AAA, where every calcium-free scan was left calcium-free, on the Siriapisith AAA/EVAR crops and on the Jamshidian 4D-CTA phases. %
\revC{} was higher than \revB{} by a paired median of 0.021 (26 better, 5 worse; $p<10^{-4}$) and than its first build by 0.009 (21/6; $p=0.004$). %

On the held-out cases we found no difference between \revC{} and \ffo{}: the pooled medians were 0.805 and 0.802, and the paired difference was $+0.006$ (20 better, 18 worse; $p=0.36$). %
The boundary and surface measures did not differ either, except NSD at 1\,mm, where \revC{} led (\cref{tab:pairs}). %
Per dataset, \revC{} led on five and trailed on AVT (\cref{tab:heldout}), with its largest lead on AortaSeg24, where \ffo{} confuses branch identities. %
The largest differences in either direction were MosMedData AAA scans, where a few calcium voxels decide the score. %
Scored on the classes present in the reference only, \ffo{} reaches a pooled median of 0.846 while \revC{} stays at 0.805, but the paired difference is $-0.015$ (17/21; $p=0.85$), so neither convention separates the two. %
The lead of \revC{} on MosMedData AAA under the primary metric comes from the calcium-specks card, which its guide tells the reviewer to check. %

The lead of \ffs{} on AVT, marked in \cref{tab:heldout}, comes from robustness rather than better segmentation. %
On two AVT scans without usable contrast for the trunk tools, every \taurus{} approach scored below 0.04, because the automated pass left an almost empty label. %
The hard cases were chosen from \revA{} failures, so they over-represent input failures that a free-form agent, which reads the CT directly, does not share, and in the main experiment \ffs{} never led a case by more than the noise floor. %
Over all held-out cases, \revC{} was higher than \ffs{} by a paired median of 0.073 (28/10, $p<0.001$). %

\section{Discussion}
\label{sec:discussion}

The free-form agent and \taurus{} win on different tasks, and the reasons are mechanical.
Where the task is one vessel and its sac, \ffo{} beat \revA{}, because it measured blood attenuation before choosing a threshold, traced the sac out to the fat and tracked the trunk from a seed, where \revA{} used fixed rules. %
None of these steps needs a language model, and once automated they carried most of the held-out gain. %
After tuning, \taurus{} still differed from the free-form agent in terracing and, on AVT, in the lumen edge, which sat at 0.34 times blood attenuation where the references sit at 0.72, and a deterministic regularization removed both differences. %
What remained were gross errors in the scored classes, and the model fixed them once the harness had found them, shown them on a card and prepared the fix.
The same reviewer that changed 12 cases in \revB{} changed 29 in \revC{}, so the model's contribution is real, but it is the judgment to accept or refuse a candidate rather than the segmentation itself. %

As a signal of label quality, the model's judgment was narrower still.
The \emph{issue} verdict separated failed from acceptable labels well only where the failure was gross.
Its AUROC was 0.82 on MosMedData AAA, where all 17 header-flipped cases received it, and 0.85 on VMR, but 0.50 to 0.63 on the other four datasets. %
Two reviews of the same case agreed little across model tiers ($\kappa = 0.06$, 54 cases) and only fairly within a tier ($\kappa = 0.34$, 24 cases).
A human who checks only the flagged cases would therefore miss most subtle failures. %
Some disagreements are probably errors in the references: auditing 26 cases outside the held-out set, we found Jamshidian 4D-CTA thrombus meshes on enhanced blood and AortaSeg24 references that leave 2--4\,cm above the valve unlabeled (other open failures are listed in \cref{tab:issues-v3}). %
Every number here treats the reference as the truth, so a label that is better than its reference counts as worse.

\paragraph{Boundary Regularization Improves Performance.}
\label{sec:res-surface}
The regularization alone made \autoB{} smoother, lowering its terracing from 0.71 to 0.50\,mm while the primary metric changed by 0.005, below the noise floor (\cref{fig:surface}). %
\revC{} brought terracing below the reference's own, 0.38 against 0.47\,mm, and raised the lumen edge level from 0.52 to 0.60, the references' level by construction (\cref{tab:surface}). %
The error maps of \cref{fig:meshes} show the change on a Siriapisith AAA/EVAR crop, where \revB{} and the first build lie inside the reference over much of the sac and \revC{} lies just outside it. %

\paragraph{Cost and the Impact of Edits.}
\label{sec:res-edits}
\label{sec:res-cost}
\ffo{} cost 4 to 33 times as much as a \revA{} review, and on the 38 held-out cases \revC{} cost a median of \$1.22 per case against \$2.71 for \ffo{}, in 40\,\% of the minutes from scan to label and a third of the turns (\cref{tab:heldout,tab:a5}). %
The saving comes from fewer and cheaper turns, because a reviewer answers a few precise questions while a free-form agent writes and debugs a pipeline. %
Most edits of the \revA{} reviewers left the primary metric unchanged, and most of the gain came from two edit commands, outer growth with the interactive segmenter on Siriapisith AAA/EVAR and the lower lumen threshold on AVT (\cref{fig:edits}). %
\ffs{} was the cheapest approach that uses a language model, at \$0.18 and 1.3 minutes per held-out case, and it lies on the cost--accuracy front (\cref{fig:pareto}).
It could serve as a cheap first pass or as a fallback where the automated pass fails. %

\paragraph{Limitations.}
\label{sec:limitations}
No expert adjudicated the outputs, the edits or the reference-disagreement notes, and we did not measure expert correction time, so every gain is measured against the references rather than against clinical acceptability. %
The free-form comparison is narrow: the free-form agents ran once per case without the trained models that \taurus{} has, and we did not run a free-form agent with the same tools or \taurus{} with Sonnet, so the gap in cost and accuracy may be narrower for an agent equipped like the harness. %
\revC{} is not a blind test: the paired controls correct for the regression to the mean of the hard cases, but not for the fact that \revB{} and \revC{} were designed around their failure modes. %
Analysis choices also shape the results: the AUROC values depend on the relative failure threshold, and the per-dataset scores on the scored region (\cref{app:data}).
Labels from \revB{} and \revC{} are still untested as training data (\cref{app:downstream}).
Dollars understate the reviews by 3--6\,\% and exclude the automated pass and the engineering of the harness, so the cost advantage holds for model cost only (\cref{app:compute}). %

\section{Conclusion}
\label{sec:conclusion}

Moving what the free-form agents did well into the automated pass of \revB{} carried most of the gain. %
\revC{}, whose cards point the reviewer at the gross errors that remained, then brought surfaces to the references' own smoothness and showed no difference from \ffo{} on the 38 held-out cases at under half of its cost. %
The model's own verdict, however, remained a weak signal of failure. %
Most of what made \taurus{} competitive therefore came from the harness, its automation and the tools that speed up the model's work, rather than from the language model itself.
Free-form agents are most useful at the start of such a project, where their transcripts show what to automate.

For annotation pipelines, the lesson is to automate what a strong agent does deterministically, to ask the model only what needs judgment, and to let it write code under checks where the tools stop.
Built this way, a harness makes relabeling public datasets to one label contract cheap, at about a dollar of model cost per case, but expert time is still needed to adjudicate the recovered labels before models can be trained on them (\cref{app:downstream}). %
\taurus{} shows that this kind of annotation can be automated, and making its labels good enough to train on is the work that remains.

\clearpage %

\section*{Data availability}
All six datasets are public, and we used each under its own licence: AortaSeg24 \citep{imran2025aortaseg24}, AVT \citep{radl2022avt}, Siriapisith AAA/EVAR \citep{siriapisith2022peerj}, the Jamshidian 4D-CTA dataset \citep{jamshidian2025fourdcta}, the Vascular Model Repository (\url{https://www.vascularmodel.com}) \citep{wilson2013vmr} and MosMedData AAA \citep{mosmeddata_aaa}.
MosMedData AAA was created by Yu.\,A.~Vasilev, M.\,R.~Kodenko, A.\,V.~Solovev, O.\,V.~Omelyanskaya, E.\,P.~Yasakova and D.\,V.~Gatin at the Research and Practical Clinical Center for Diagnostics and Telemedicine Technologies of the Moscow Health Care Department (Copyright \textcopyright{} 2020), is available at \url{https://mosmed.ai/datasets/} and is licensed under CC BY-NC-ND 3.0 (\url{https://creativecommons.org/licenses/by-nc-nd/3.0/}).
We report scores on MosMedData AAA but show no images from it.
VMR image data in \cref{tab:datasets,fig:overlays} are \textcopyright{} Stanford University, the Regents of the University of California, Open Source Medical Software Corporation, and other parties.
We do not redistribute any of the datasets, and we do not release labels derived from MosMedData AAA or VMR.
The in-house scan used only during development is not public.
The harness code, the other derived labels and the result files are available from the authors on reasonable request, subject to the dataset licences.

\section*{Acknowledgements}
The VMR data used herein was provided in whole or in part with Federal funds from the National Library of Medicine under Grant No.~R01LM013120, and the National Heart, Lung, and Blood Institute, National Institutes of Health, Department of Health and Human Services, under Contract No.~HHSN268201100035C.

\section*{Use of AI tools}
Claude language models (Anthropic) are the subject of this study and were also used as tools in it.
The authors checked the code, the numbers and the text and take responsibility for the content.

\bibliographystyle{unsrtnat}
\bibliography{refs}

\clearpage
\appendix
\renewcommand{\topfraction}{0.85}
\renewcommand{\bottomfraction}{0.6}
\renewcommand{\textfraction}{0.1}
\renewcommand{\floatpagefraction}{0.75}
\setcounter{topnumber}{3}
\setcounter{bottomnumber}{1}
\setcounter{totalnumber}{4}
\iffullappendix
  \input{sections/A_appendix_full}
\else
  
\section{Datasets, sampling and scored region}
\label{app:data}

\emph{Datasets.}
We used six public datasets with 939 volumes (\cref{tab:datasets}): 100 whole-body angiograms of type B dissections from the AortaSeg24 challenge dataset with 23 expert vessel ids \citep{imran2025aortaseg24}, of which we score the 14 abdominal ids because our thoracic zones are unverified; 520 aneurysm and endograft crops of $256\times256\times64$ voxels (about 80\,mm across) from the Siriapisith pre- and post-operative AAA CTA set (Siriapisith AAA/EVAR), labeled with the outer envelope \citep{siriapisith2022peerj}; 98 scans of MosMedData abdominal aorta CTA (MosMedData AAA) with lumen and calcium \citep{mosmeddata_aaa,kodenko2025dataset}; 55 scans of the Aortic Vessel Tree (AVT) collection from three sources with a binary lumen tree \citep{radl2022avt}; 17 models of the Vascular Model Repository (VMR) with a lofted lumen per named vessel \citep{wilson2013vmr}; and 149 cardiac phases of 20 patients from the 4D-CTA image and geometry dataset of abdominal aortic aneurysms (Jamshidian 4D-CTA) with thrombus and sac meshes \citep{jamshidian2025fourdcta}. %
Development also used one in-house scan, AVT Rider R6 with a four-class tissue reference drawn by an expert reviewer.
It was among the free-form cases and the pilot of \revC{}, but because it is a single case we leave it out of all results. %

The starting vessels came from our nnU-Net \citep{isensee2021nnunet} vessel model with a branch refiner on AortaSeg24 (out-of-fold, since it was trained there), MosMedData AAA, AVT and VMR, and from TotalSegmentator 2.13 \citep{wasserthal2023totalseg} on Siriapisith AAA/EVAR and the Jamshidian 4D-CTA phases. %
The 27 vessel labels of the label contract are the ten aortic zones 0--9, the brachiocephalic trunk, left common carotid, left subclavian, celiac trunk, superior and inferior mesenteric arteries, the right and left renal, accessory renal, common iliac, internal iliac and external iliac arteries, and one label for any other aortic branch. %

\emph{Sampling.}
Samples were drawn by a salted SHA-1 hash of the case key within strata (for example MosMedData AAA by pathology, Jamshidian by patient) and frozen before any run. %
\revA{} labels exist for 438 cases: all AortaSeg24 and MosMedData AAA cases, 160 Siriapisith AAA/EVAR crops, 24 AVT cases, all VMR models and 39 Jamshidian phases (two per patient). %
\autoA{} ran on all 939 volumes; it produced no output on 30 (28 Siriapisith AAA/EVAR crops, one Jamshidian 4D-CTA phase, one AVT case) because the proposal found no aorta, and only the phase was in a review sample. %
Identity accuracy (AortaSeg24) is the share of voxels labeled as vessel in both maps that carry the same id. %

\emph{Scored region.}
The study plan described the scored region as the human label restricted to the vessels and zones of our label contract and dilated by 5\,mm. %
The scorer, written the same day and before the first run, implements it per dataset instead: overlap on AortaSeg24 and Siriapisith AAA/EVAR is scored on the whole volume (the AortaSeg24 confusion matrix within 3\,mm of either label), on MosMedData AAA within the axial extent of the human label and 15\,mm of it, on VMR within the axial extent of the human label and 10\,mm of either label, on AVT within the axial extent of our trunk and iliac labels and 10\,mm of either label, and on Jamshidian 4D-CTA within the axial extent of the reference sac. %
No scoring rule changed after the runs began except the VMR id merge: VMR labels the whole aorta with one id, so we merged our trunk zones before scoring, a change made after the first runs and applied to every approach. %

\FloatBarrier
\section{Paired statistics}
\label{app:tables}
\label{app:main-results}

This appendix gives the paired comparisons behind the main-text results: all main-experiment pairs in \cref{tab:app-pairs}, the held-out medians in \cref{tab:a5} and all held-out pairs in \cref{fig:heldout-pairs}. %

\begin{table}[h]
  \centering
  \caption{Paired comparisons behind the main-text Results: median difference of the first approach minus the second, bootstrap 95\,\% CI where computed, cases better and worse by more than 0.005 for the first approach, and Wilcoxon $p$.
  The two base-review rows of the main experiment are Holm-adjusted, all others unadjusted.
  \taurus{} approaches are named \revA{}, \revB{} and \revC{}, with (auto) marking the automated pass alone.
  For MSD, HD95 and terracing, lower is better.}
  \label{tab:pairs}
  \footnotesize
  \setlength{\tabcolsep}{4pt}
  \begin{tabular}{@{}l l l r l c l@{}}
    \toprule
    Comparison & Cases & Metric & \makecell{Median\\$\Delta$} & 95\,\% CI & \makecell{Better/\\worse} & $p$ \\
    \midrule
    \multicolumn{7}{@{}l}{\textit{Main experiment}} \\
    base $-$ base (auto) & Siriapisith AAA/EVAR, 160 & outer Dice & $+0.013$ & 0.000 to 0.025 & 92/15 & $1.5\times10^{-14}$ \\ %
    base $-$ base (auto) & AVT, 24 & tree Dice & $+0.022$ & 0.001 to 0.049 & 21/1 & 0.0024 \\ %
    hybrid $-$ \ffo{} & free-form, 30 & primary metric & $+0.027$ & -- & 21/6 & 0.001 \\ %
     & & NSD at 1\,mm & $+0.095$ & -- & 23/6 & 0.0003 \\ %
     & & MSD (mm) & $-0.53$ & -- & 24/5 & 0.001 \\ %
     & & terracing (mm) & $-0.002$ & -- & 13/12 & 0.65 \\ %
    \midrule
    \multicolumn{7}{@{}l}{\textit{Held-out cases}} \\
    tuned (auto) $-$ base (auto) & held-out, 38 & primary metric & $+0.258$ & 0.076 to 0.389 & 32/2 & $8.8\times10^{-7}$ \\ %
    tuned $-$ tuned (auto) & held-out, 38 & primary metric & $+0.000$ & 0.000 to 0.001 & 11/1 & 0.019 \\ %
    base $-$ base (auto) & held-out, 38 & primary metric & $+0.000$ & 0.000 to 0.002 & 11/1 & 0.0032 \\ %
    hybrid (auto) $-$ tuned (auto) & held-out, 38 & primary metric & $+0.005$ & 0.003 to 0.009 & 19/2 & $3.0\times10^{-5}$ \\ %
     & & terracing (mm) & $-0.168$ & -- & 27/4 & $2.2\times10^{-7}$ \\ %
    hybrid $-$ hybrid (auto) & held-out, 38 & primary metric & $+0.025$ & 0.004 to 0.108 & 24/5 & 0.0001 \\ %
     & & MSD (mm) & $-0.33$ & -- & 24/4 & 0.0004 \\ %
     & & HD95 (mm) & $-0.09$ & -- & 19/7 & 0.004 \\ %
    hybrid $-$ tuned & held-out, 38 & primary metric & $+0.021$ & 0.006 to 0.041 & 26/5 & $1.9\times10^{-5}$ \\ %
     & & NSD at 1\,mm & $+0.017$ & -- & 25/9 & 0.0015 \\ %
     & & terracing (mm) & $-0.155$ & -- & 33/2 & $6.0\times10^{-9}$ \\ %
     & & HD95 (mm) & $-0.348$ & -- & 21/14 & 0.096 \\ %
    hybrid $-$ first hybrid build & held-out, 38 & primary metric & $+0.009$ & 0.001 to 0.017 & 21/6 & 0.0037 \\ %
    hybrid $-$ \ffs{} & held-out, 38 & primary metric & $+0.073$ & 0.039 to 0.167 & 28/10 & 0.0009 \\ %
    hybrid $-$ \ffo{} & held-out, 38 & primary metric & $+0.006$ & $-0.022$ to 0.045 & 20/18 & 0.36 \\ %
     & & NSD at 1\,mm & $+0.046$ & -- & 23/14 & 0.045 \\ %
     & & MSD (mm) & $-0.11$ & -- & 19/17 & 0.36 \\ %
     & & terracing (mm) & $-0.002$ & -- & 17/16 & 0.85 \\ %
    tuned $-$ \ffo{} & held-out, 38 & primary metric & $-0.023$ & $-0.049$ to $-0.007$ & 11/26 & 0.067 \\ %
    \bottomrule
  \end{tabular}
\end{table}

\begin{table}[h]
  \centering
  \caption{Paired comparisons of the primary metric in the main experiment: median difference, bootstrap 95\,\% interval, Wilcoxon $p$, Holm-adjusted $p$ and the number of cases better/worse for the first approach.
  \src{\texttt{results/tables/T2\_pairs.md}}}
  \label{tab:app-pairs}
  \scriptsize
  \setlength{\tabcolsep}{4pt}
  \begin{tabular}{@{}llrrlrrl@{}}
\toprule
Comparison & Dataset & $n$ & median $\Delta$ & 95\,\% CI & $p$ & $p$ (Holm) & better/worse \\
\midrule
\revA{} $-$ \autoA{} & AortaSeg24 & 100 & +0.000 & [+0.000, +0.000] & 4.83e-05 & 0.00246 & 45/15 \\
\revA{} $-$ \autoA{} & Siriapisith AAA/EVAR & 160 & +0.013 & [+0.000, +0.025] & 2.62e-16 & 1.49e-14 & 92/15 \\
\revA{} $-$ \autoA{} & MosMedData AAA & 98 & +0.000 & [+0.000, +0.000] & 0.817 & 1 & 18/16 \\
\revA{} $-$ \autoA{} & AVT & 24 & +0.022 & [+0.001, +0.049] & 4.61e-05 & 0.0024 & 21/1 \\
\revA{} $-$ \autoA{} & VMR & 17 & +0.000 & [+0.000, +0.002] & 0.0754 & 1 & 8/3 \\
\revA{} $-$ \autoA{} & Jamshidian 4D-CTA & 39 & +0.000 & [+0.000, +0.000] & 0.291 & 1 & 10/12 \\
\addlinespace[2pt]
Crossover $-$ \autoA{} & AortaSeg24 & 10 & +0.000 & [+0.000, +0.000] & 1 & 1 & 0/1 \\
Crossover $-$ \autoA{} & Siriapisith AAA/EVAR & 10 & +0.216 & [+0.099, +0.487] & 0.00195 & 0.0977 & 10/0 \\
Crossover $-$ \autoA{} & MosMedData AAA & 10 & +0.000 & [+0.000, +0.000] & 0.5 & 1 & 2/0 \\
Crossover $-$ \autoA{} & AVT & 8 & +0.000 & [+0.000, +0.000] & — & — & 0/0 \\
Crossover $-$ \autoA{} & VMR & 6 & -0.000 & [-0.025, +0.000] & 0.25 & 1 & 0/3 \\
Crossover $-$ \autoA{} & Jamshidian 4D-CTA & 10 & +0.000 & [-0.002, +0.019] & 0.875 & 1 & 2/2 \\
\addlinespace[2pt]
Repeat $-$ \autoA{} & AortaSeg24 & 5 & +0.005 & [+0.000, +0.021] & 0.25 & 1 & 3/0 \\
Repeat $-$ \autoA{} & Siriapisith AAA/EVAR & 5 & +0.021 & [+0.000, +0.220] & 0.125 & 1 & 4/0 \\
Repeat $-$ \autoA{} & MosMedData AAA & 5 & +0.000 & [+0.000, +0.044] & 1 & 1 & 1/0 \\
Repeat $-$ \autoA{} & AVT & 3 & +0.039 & [+0.000, +0.807] & 0.25 & 1 & 3/0 \\
Repeat $-$ \autoA{} & VMR & 3 & +0.002 & [+0.000, +0.010] & 0.5 & 1 & 2/0 \\
Repeat $-$ \autoA{} & Jamshidian 4D-CTA & 3 & +0.229 & [+0.000, +0.402] & 0.5 & 1 & 2/0 \\
\addlinespace[2pt]
Crossover $-$ \revA{} & AortaSeg24 & 10 & -0.000 & [-0.004, +0.000] & 0.0625 & 1 & 0/5 \\
Crossover $-$ \revA{} & Siriapisith AAA/EVAR & 10 & +0.110 & [+0.054, +0.243] & 0.00195 & 0.0977 & 10/0 \\
Crossover $-$ \revA{} & MosMedData AAA & 10 & +0.000 & [-0.000, +0.000] & 1 & 1 & 2/3 \\
Crossover $-$ \revA{} & AVT & 8 & -0.019 & [-0.045, -0.001] & 0.00781 & 0.375 & 0/8 \\
Crossover $-$ \revA{} & VMR & 6 & -0.000 & [-0.032, +0.003] & 0.625 & 1 & 1/3 \\
Crossover $-$ \revA{} & Jamshidian 4D-CTA & 10 & -0.010 & [-0.500, +0.043] & 0.312 & 1 & 3/5 \\
\addlinespace[2pt]
Repeat $-$ \revA{} & AortaSeg24 & 5 & +0.001 & [-0.010, +0.005] & 0.875 & 1 & 3/1 \\
Repeat $-$ \revA{} & Siriapisith AAA/EVAR & 5 & +0.000 & [-0.109, +0.041] & 1 & 1 & 2/2 \\
Repeat $-$ \revA{} & MosMedData AAA & 5 & +0.000 & [+0.000, +0.044] & 1 & 1 & 1/0 \\
Repeat $-$ \revA{} & AVT & 3 & +0.002 & [-0.000, +0.005] & 0.5 & 1 & 2/1 \\
Repeat $-$ \revA{} & VMR & 3 & +0.000 & [-0.000, +0.000] & 1 & 1 & 1/1 \\
Repeat $-$ \revA{} & Jamshidian 4D-CTA & 3 & -0.030 & [-0.042, +0.000] & 0.5 & 1 & 0/2 \\
\addlinespace[2pt]
\revA{} $-$ \ffo{} & AortaSeg24 & 5 & +0.080 & [+0.028, +0.184] & 0.0625 & 1 & 5/0 \\
\revA{} $-$ \ffo{} & Siriapisith AAA/EVAR & 6 & -0.157 & [-0.462, -0.125] & 0.0312 & 1 & 0/6 \\
\revA{} $-$ \ffo{} & MosMedData AAA & 5 & -0.081 & [-0.807, +0.039] & 0.188 & 1 & 1/4 \\
\revA{} $-$ \ffo{} & AVT & 5 & -0.063 & [-0.100, -0.012] & 0.0625 & 1 & 0/5 \\
\revA{} $-$ \ffo{} & VMR & 4 & +0.003 & [-0.923, +0.021] & 0.875 & 1 & 3/1 \\
\revA{} $-$ \ffo{} & Jamshidian 4D-CTA & 5 & -0.223 & [-0.704, +0.021] & 0.312 & 1 & 2/3 \\
\addlinespace[2pt]
\revA{} $-$ \ffs{} & AortaSeg24 & 5 & +0.381 & [+0.093, +0.667] & 0.0625 & 1 & 5/0 \\
\revA{} $-$ \ffs{} & Siriapisith AAA/EVAR & 6 & -0.132 & [-0.395, -0.055] & 0.0312 & 1 & 0/6 \\
\revA{} $-$ \ffs{} & MosMedData AAA & 5 & +0.010 & [-0.465, +0.158] & 0.812 & 1 & 3/2 \\
\revA{} $-$ \ffs{} & AVT & 5 & +0.172 & [-0.062, +0.275] & 0.125 & 1 & 4/1 \\
\revA{} $-$ \ffs{} & VMR & 4 & +0.355 & [-0.915, +0.447] & 0.875 & 1 & 3/1 \\
\revA{} $-$ \ffs{} & Jamshidian 4D-CTA & 5 & +0.004 & [-0.568, +0.260] & 1 & 1 & 3/2 \\
\addlinespace[2pt]
\ffo{} $-$ \ffs{} & AortaSeg24 & 5 & +0.300 & [+0.065, +0.594] & 0.0625 & 1 & 5/0 \\
\ffo{} $-$ \ffs{} & Siriapisith AAA/EVAR & 6 & +0.041 & [-0.017, +0.137] & 0.156 & 1 & 5/1 \\
\ffo{} $-$ \ffs{} & MosMedData AAA & 5 & +0.118 & [+0.019, +0.343] & 0.0625 & 1 & 5/0 \\
\ffo{} $-$ \ffs{} & AVT & 5 & +0.232 & [+0.023, +0.338] & 0.0625 & 1 & 5/0 \\
\ffo{} $-$ \ffs{} & VMR & 4 & +0.344 & [+0.009, +0.443] & 0.125 & 1 & 4/0 \\
\ffo{} $-$ \ffs{} & Jamshidian 4D-CTA & 5 & +0.239 & [+0.000, +0.473] & 0.125 & 1 & 4/0 \\
\addlinespace[2pt]
\revA{} $-$ \bel{} & AortaSeg24 & 30 & -0.135 & [-0.140, -0.122] & 2.56e-06 & 0.000138 & 0/29 \\
\revA{} $-$ \bel{} & MosMedData AAA & 39 & +0.191 & [-0.003, +0.260] & 0.0349 & 1 & 22/16 \\
\revA{} $-$ \bel{} & AVT & 24 & +0.011 & [+0.002, +0.019] & 0.034 & 1 & 18/6 \\
\revA{} $-$ \bel{} & VMR & 17 & +0.014 & [+0.000, +0.023] & 0.0267 & 1 & 13/4 \\
\addlinespace[2pt]
\revA{} $-$ \tsg{} & Siriapisith AAA/EVAR & 60 & +0.598 & [+0.544, +0.714] & 1.8e-11 & 1.01e-09 & 59/1 \\
\addlinespace[2pt]
\autoA{} $-$ \bel{} & AortaSeg24 & 30 & -0.140 & [-0.147, -0.124] & 2.56e-06 & 0.000138 & 0/29 \\
\autoA{} $-$ \bel{} & MosMedData AAA & 39 & +0.000 & [-0.011, +0.243] & 0.162 & 1 & 19/19 \\
\autoA{} $-$ \bel{} & AVT & 24 & -0.018 & [-0.047, +0.010] & 0.0604 & 1 & 10/14 \\
\autoA{} $-$ \bel{} & VMR & 17 & +0.009 & [-0.002, +0.022] & 0.0984 & 1 & 12/5 \\
\addlinespace[2pt]
\autoA{} $-$ \tsg{} & Siriapisith AAA/EVAR & 60 & +0.502 & [+0.439, +0.562] & 1.99e-11 & 1.1e-09 & 58/2 \\
\bottomrule
\end{tabular}

\end{table}

\begin{table*}[t]
  \centering
  \caption{Full held-out results on the 38 evaluation cases (26 hard, 12 typical), the long form of \cref{tab:heldout}: median primary metric per dataset and over all, hard and typical cases (each dataset in its own metric, so the pooled columns are descriptive), and LLM cost, minutes from scan to label (vessel proposal and automated pass, plus the review for the reviewed approaches) and turns per case; $\pm$ half the interquartile range beneath each median.
  Bold marks the highest median of the column, and asterisks mark medians that rest on two cases where an approach left an almost empty label and on the calcium scoring convention.
  \Cref{fig:heldout-pairs} draws the paired differences, and \cref{tab:pairs} in the main text gives those quoted there.
  \autoA{} and \revA{} form the paired control, re-run and re-reviewed on the same cases; \revB{} starts from the frozen \autoB{} labels and \revC{} from its own automated pass, regenerated from the same proposals; \ffo{} ran on all 38 cases. %
  \revC{} was designed after the results of \revB{} and of its own first build were seen, so its comparison is not blind.
  \src{\texttt{results/tables/T8\_A5.md}, \texttt{T11\_v31\_eval.md} and \texttt{T11\_v31\_eval\_pairs.md}}}
  \label{tab:a5}
  \scriptsize\renewcommand{\footnotesize}{\scriptsize}\renewcommand{\arraystretch}{0.92}%
  \begin{table*}[t]
\centering
\footnotesize
\caption{Held-out evaluation set (38 cases): median primary metric per dataset and over all, hard and typical cases, and cost, minutes and turns per case; each cell gives the median and, beneath it, $\pm$ half the interquartile range. Superscript (n) = cases when an approach ran on fewer than the column (Free-form Opus: 38 hard cases). Bold = the highest median among the approaches run on every case of the column. Each dataset keeps its own primary metric (Table~\ref{tab:datasets}). Cost: list-price equivalents of the review transcripts (reviews) and the CLI's own cost (free-form); the automated passes run on CPU with no LLM cost. Minutes run from scan to label: the vessel proposal on the GPU and the automated pass on CPU (measured on one workstation, one case at a time), plus the review for the reviewed approaches; free-form, the run's wall time.}
\label{tab:heldout}
\setlength{\tabcolsep}{0.5pt}
\begin{tabular*}{\textwidth}{@{\extracolsep{\fill}}lcccccccccccc@{}}
\toprule
 & \multicolumn{6}{c}{Primary metric per dataset} & \multicolumn{3}{c}{All held-out cases / stratum} & \multicolumn{3}{c}{Cost per case} \\
\cmidrule(lr){2-7}\cmidrule(lr){8-10}\cmidrule(l){11-13}
Approach & \makecell{AortaSeg24\\[-1pt](4)} & \makecell{Siriapisith\\AAA/EVAR\\[-1pt](12)} & \makecell{MosMedData\\AAA\\[-1pt](8)} & \makecell{AVT\\[-1pt](4)} & \makecell{VMR\\[-1pt](4)} & \makecell{Jamshidian\\4D-CTA\\[-1pt](6)} & \makecell{all\\[-1pt](38)} & \makecell{hard\\[-1pt](26)} & \makecell{typical\\[-1pt](12)} & \$ & min & turns \\
\midrule
\makecell[l]{Taurus-\\base\\(auto)} & \makecell{0.572\\[-2pt]\scriptsize $\pm$0.072} & \makecell{0.496\\[-2pt]\scriptsize $\pm$0.096} & \makecell{0.021\\[-2pt]\scriptsize $\pm$0.071} & \makecell{0.027\\[-2pt]\scriptsize $\pm$0.108} & \makecell{0.490\\[-2pt]\scriptsize $\pm$0.043} & \makecell{0.002\\[-2pt]\scriptsize $\pm$0.005} & \makecell{0.321\\[-2pt]\scriptsize $\pm$0.259} & \makecell{0.109\\[-2pt]\scriptsize $\pm$0.225} & \makecell{0.573\\[-2pt]\scriptsize $\pm$0.151} & -- & \makecell{0.7\\[-2pt]\scriptsize $\pm$1.1} & -- \\ \addlinespace[1pt]
\makecell[l]{Taurus-\\base} & \makecell{0.579\\[-2pt]\scriptsize $\pm$0.076} & \makecell{0.496\\[-2pt]\scriptsize $\pm$0.096} & \makecell{0.024\\[-2pt]\scriptsize $\pm$0.076} & \makecell{0.432\\[-2pt]\scriptsize $\pm$0.405} & \makecell{\textbf{0.492}\\[-2pt]\scriptsize $\pm$0.037} & \makecell{0.004\\[-2pt]\scriptsize $\pm$0.004} & \makecell{0.410\\[-2pt]\scriptsize $\pm$0.260} & \makecell{0.144\\[-2pt]\scriptsize $\pm$0.222} & \makecell{0.640\\[-2pt]\scriptsize $\pm$0.146} & \makecell{0.20\\[-2pt]\scriptsize $\pm$0.44} & \makecell{1.5\\[-2pt]\scriptsize $\pm$4.2} & \makecell{8\\[-2pt]\scriptsize $\pm$8} \\ \addlinespace[1pt]
\makecell[l]{Taurus-\\tuned\\(auto)} & \makecell{0.732\\[-2pt]\scriptsize $\pm$0.031} & \makecell{0.783\\[-2pt]\scriptsize $\pm$0.130} & \makecell{0.462\\[-2pt]\scriptsize $\pm$0.089} & \makecell{0.437\\[-2pt]\scriptsize $\pm$0.406} & \makecell{0.464\\[-2pt]\scriptsize $\pm$0.048} & \makecell{0.534\\[-2pt]\scriptsize $\pm$0.186} & \makecell{0.620\\[-2pt]\scriptsize $\pm$0.159} & \makecell{0.598\\[-2pt]\scriptsize $\pm$0.145} & \makecell{0.716\\[-2pt]\scriptsize $\pm$0.195} & -- & \makecell{1.0\\[-2pt]\scriptsize $\pm$1.1} & -- \\ \addlinespace[1pt]
\makecell[l]{Taurus-\\tuned} & \makecell{0.732\\[-2pt]\scriptsize $\pm$0.032} & \makecell{0.871\\[-2pt]\scriptsize $\pm$0.037} & \makecell{0.462\\[-2pt]\scriptsize $\pm$0.104} & \makecell{0.437\\[-2pt]\scriptsize $\pm$0.406} & \makecell{0.466\\[-2pt]\scriptsize $\pm$0.064} & \makecell{0.572\\[-2pt]\scriptsize $\pm$0.222} & \makecell{0.704\\[-2pt]\scriptsize $\pm$0.202} & \makecell{0.698\\[-2pt]\scriptsize $\pm$0.199} & \makecell{0.716\\[-2pt]\scriptsize $\pm$0.196} & \makecell{0.75\\[-2pt]\scriptsize $\pm$0.23} & \makecell{6.4\\[-2pt]\scriptsize $\pm$4.7} & \makecell{16\\[-2pt]\scriptsize $\pm$5} \\ \addlinespace[1pt]
\makecell[l]{Taurus-\\hybrid\\(auto)} & \makecell{\textbf{0.734}\\[-2pt]\scriptsize $\pm$0.031} & \makecell{0.791\\[-2pt]\scriptsize $\pm$0.135} & \makecell{0.468\\[-2pt]\scriptsize $\pm$0.105} & \makecell{0.454\\[-2pt]\scriptsize $\pm$0.428} & \makecell{0.468\\[-2pt]\scriptsize $\pm$0.047} & \makecell{0.587\\[-2pt]\scriptsize $\pm$0.180} & \makecell{0.647\\[-2pt]\scriptsize $\pm$0.154} & \makecell{0.619\\[-2pt]\scriptsize $\pm$0.137} & \makecell{0.725\\[-2pt]\scriptsize $\pm$0.202} & -- & \makecell{1.1\\[-2pt]\scriptsize $\pm$1.3} & -- \\ \addlinespace[1pt]
\makecell[l]{Taurus-\\hybrid} & \makecell{0.733\\[-2pt]\scriptsize $\pm$0.026} & \makecell{\textbf{0.897}\\[-2pt]\scriptsize $\pm$0.022} & \makecell{\textbf{0.853}\\[-2pt]\scriptsize $\pm$0.104} & \makecell{0.442\\[-2pt]\scriptsize $\pm$0.435} & \makecell{0.483\\[-2pt]\scriptsize $\pm$0.073} & \makecell{\textbf{0.732}\\[-2pt]\scriptsize $\pm$0.043} & \makecell{\textbf{0.805}\\[-2pt]\scriptsize $\pm$0.114} & \makecell{0.818\\[-2pt]\scriptsize $\pm$0.110} & \makecell{\textbf{0.762}\\[-2pt]\scriptsize $\pm$0.115} & \makecell{1.22\\[-2pt]\scriptsize $\pm$0.45} & \makecell{6.4\\[-2pt]\scriptsize $\pm$5.0} & \makecell{23\\[-2pt]\scriptsize $\pm$10} \\ \addlinespace[1pt]
\makecell[l]{Free-form\\Sonnet} & \makecell{0.307\\[-2pt]\scriptsize $\pm$0.071} & \makecell{0.802\\[-2pt]\scriptsize $\pm$0.045} & \makecell{0.773$^{*}$\\[-2pt]\scriptsize $\pm$0.177} & \makecell{\textbf{0.823}$^{*}$\\[-2pt]\scriptsize $\pm$0.099} & \makecell{0.226\\[-2pt]\scriptsize $\pm$0.126} & \makecell{0.430\\[-2pt]\scriptsize $\pm$0.262} & \makecell{0.677\\[-2pt]\scriptsize $\pm$0.195} & \makecell{0.714\\[-2pt]\scriptsize $\pm$0.150} & \makecell{0.441\\[-2pt]\scriptsize $\pm$0.372} & \makecell{0.18\\[-2pt]\scriptsize $\pm$0.04} & \makecell{1.3\\[-2pt]\scriptsize $\pm$0.8} & \makecell{13\\[-2pt]\scriptsize $\pm$4} \\ \addlinespace[1pt]
\makecell[l]{Free-form\\Opus} & \makecell{0.629\\[-2pt]\scriptsize $\pm$0.033} & \makecell{0.892\\[-2pt]\scriptsize $\pm$0.021} & \makecell{0.837\\[-2pt]\scriptsize $\pm$0.203} & \makecell{0.466$^{*}$\\[-2pt]\scriptsize $\pm$0.432} & \makecell{0.446\\[-2pt]\scriptsize $\pm$0.035} & \makecell{0.715\\[-2pt]\scriptsize $\pm$0.052} & \makecell{0.802\\[-2pt]\scriptsize $\pm$0.196} & \makecell{\textbf{0.829}\\[-2pt]\scriptsize $\pm$0.165} & \makecell{0.665\\[-2pt]\scriptsize $\pm$0.208} & \makecell{2.71\\[-2pt]\scriptsize $\pm$1.51} & \makecell{16.1\\[-2pt]\scriptsize $\pm$8.2} & \makecell{70\\[-2pt]\scriptsize $\pm$25} \\ \addlinespace[1pt]
\bottomrule
\end{tabular*}
\end{table*}
\end{table*}

\begin{figure}[h]
  \centering
  \figorbox{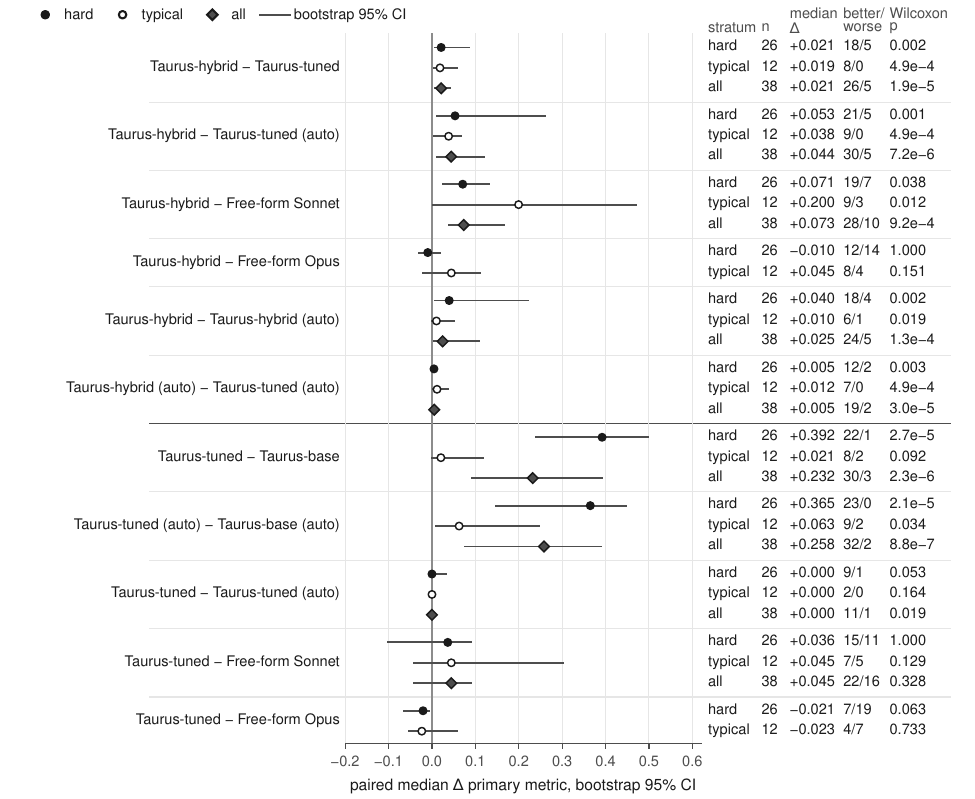}{\linewidth}{6.0cm}
  \caption{All paired comparisons of the primary metric on the held-out cases, per stratum (hard, typical, all): median difference with bootstrap 95\,\% interval, cases better/worse and unadjusted Wilcoxon $p$.
  \src{\texttt{results/tables/T8\_A5\_pairs.csv} and \texttt{T11\_v31\_eval\_pairs.md}}}
  \label{fig:heldout-pairs}
\end{figure}

\FloatBarrier
\section{Surface metrics}
\label{app:surface}

Terracing is the mean deviation of each axial slice's tracked 2D component from the mean of its two neighbors, in equivalent radius and centroid; roughness is the mean distance of the voxel surface from a Gaussian-smoothed copy ($\sigma=1.5$\,mm); the signed edge offset is the mean in-plane distance from an approach's boundary to the reference's edge on slices where both exist (clipped at $\pm5$\,mm, positive outside); and the lumen edge level is the mean of the median attenuation just inside and just outside the boundary on a CT smoothed in-plane at 0.7\,mm, divided by blood attenuation (the median of the reference structure eroded twice). %
Distances are undefined for a class that one side leaves empty, where NSD is 0, and the surface scorer reproduces the primary scorer's Dice and volumes on all 2,111 class rows it was checked on. %

\begin{table}[h]
  \centering
  \caption{Boundary and surface quality on the 38 held-out cases against the human reference.
  Top and middle: median [IQR] per approach of Dice, NSD at 1 and 2\,mm, mean surface distance (MSD), HD95, signed edge offset, lumen edge level, terracing and roughness, with the reference's own edge level, terracing and roughness ($n$ in brackets where a metric is undefined on some cases).
  Bottom: per-dataset medians of NSD at 1\,mm and MSD.
  Per case each value is the mean over the dataset's primary-metric classes; pooled medians are descriptive.
  Paired tests: \cref{tab:pairs}.
  \src{\texttt{results/tables/T9\_surface.md} (block A) and \texttt{T11\_v31\_eval.md}}}
  \label{tab:surface}
  \small
  \begin{table}[t]
\centering
\scriptsize
\caption{Boundary and surface quality against the human reference on the 38 held-out cases, median [IQR] over the cases pooled across datasets (descriptive; n in brackets when a metric is undefined on some cases), and per-dataset medians of NSD 1 mm and MSD (bottom). Per case, each value is the mean over the dataset's primary-metric classes. NSD = share of boundary within 1 / 2 mm of the other surface; MSD = symmetric mean surface distance; HD95 = 95th-percentile Hausdorff distance (distances undefined when one side leaves a class empty); edge offset = mean signed in-plane distance of the boundary to the reference's (+ = outside); edge level = where the label puts the contrast boundary, in units of the case's blood HU (no lumen reference on the crops); terracing = slice-to-slice jitter of each axial section's equivalent radius and centroid; roughness = mean distance of the voxel surface from a smoothed copy. Free-form Opus ran on 38 cases. Paired tests: Table~\ref{tab:a5} and Figure~\ref{fig:surface}.}
\label{tab:surface}
\setlength{\tabcolsep}{3pt}
\begin{tabular}{@{}lccccc@{}}
\toprule
Approach (cases) & Dice & NSD 1 mm & NSD 2 mm & MSD (mm) & HD95 (mm) \\
\midrule
\makecell[l]{Taurus-\\base (auto) (38)} & \makecell{0.321\\[-1pt]\scriptsize[0.018, 0.536]} & \makecell{0.118\\[-1pt]\scriptsize[0.014, 0.366]} & \makecell{0.233\\[-1pt]\scriptsize[0.040, 0.490]} & \makecell{12.73\\[-1pt]\scriptsize[6.79, 18.68] (37)} & \makecell{29.9\\[-1pt]\scriptsize[22.0, 54.5] (37)} \\ \addlinespace[1.5pt]
\makecell[l]{Taurus-\\base (38)} & \makecell{0.410\\[-1pt]\scriptsize[0.021, 0.542]} & \makecell{0.153\\[-1pt]\scriptsize[0.026, 0.432]} & \makecell{0.290\\[-1pt]\scriptsize[0.044, 0.575]} & \makecell{11.43\\[-1pt]\scriptsize[5.84, 18.51] (37)} & \makecell{29.2\\[-1pt]\scriptsize[19.8, 44.5] (37)} \\ \addlinespace[1.5pt]
\makecell[l]{Taurus-\\tuned (auto) (38)} & \makecell{0.620\\[-1pt]\scriptsize[0.451, 0.769]} & \makecell{0.393\\[-1pt]\scriptsize[0.254, 0.560]} & \makecell{0.588\\[-1pt]\scriptsize[0.458, 0.747]} & \makecell{2.59\\[-1pt]\scriptsize[1.67, 6.51] (37)} & \makecell{14.4\\[-1pt]\scriptsize[6.3, 23.5] (37)} \\ \addlinespace[1.5pt]
\makecell[l]{Taurus-\\tuned (38)} & \makecell{0.704\\[-1pt]\scriptsize[0.450, 0.854]} & \makecell{0.431\\[-1pt]\scriptsize[0.276, 0.566]} & \makecell{0.662\\[-1pt]\scriptsize[0.477, 0.768]} & \makecell{2.00\\[-1pt]\scriptsize[1.43, 5.61] (37)} & \makecell{9.8\\[-1pt]\scriptsize[5.5, 20.2] (37)} \\ \addlinespace[1.5pt]
\makecell[l]{Taurus-\\hybrid (auto) (38)} & \makecell{0.647\\[-1pt]\scriptsize[0.466, 0.773]} & \makecell{0.402\\[-1pt]\scriptsize[0.255, 0.577]} & \makecell{0.582\\[-1pt]\scriptsize[0.457, 0.756]} & \makecell{2.42\\[-1pt]\scriptsize[1.64, 6.26] (37)} & \makecell{14.1\\[-1pt]\scriptsize[7.1, 24.1] (37)} \\ \addlinespace[1.5pt]
\makecell[l]{Taurus-\\hybrid (38)} & \makecell{0.805\\[-1pt]\scriptsize[0.670, 0.899]} & \makecell{0.494\\[-1pt]\scriptsize[0.371, 0.702]} & \makecell{0.761\\[-1pt]\scriptsize[0.629, 0.860]} & \makecell{1.58\\[-1pt]\scriptsize[1.07, 3.20]} & \makecell{8.0\\[-1pt]\scriptsize[5.0, 14.0]} \\ \addlinespace[1.5pt]
\makecell[l]{Free-form\\Sonnet (38)} & \makecell{0.677\\[-1pt]\scriptsize[0.417, 0.808]} & \makecell{0.301\\[-1pt]\scriptsize[0.175, 0.461]} & \makecell{0.489\\[-1pt]\scriptsize[0.351, 0.679]} & \makecell{3.19\\[-1pt]\scriptsize[1.45, 7.76] (36)} & \makecell{14.8\\[-1pt]\scriptsize[6.9, 28.1] (36)} \\ \addlinespace[1.5pt]
\makecell[l]{Free-form\\Opus (38)} & \makecell{0.802\\[-1pt]\scriptsize[0.497, 0.889]} & \makecell{0.405\\[-1pt]\scriptsize[0.260, 0.600]} & \makecell{0.638\\[-1pt]\scriptsize[0.504, 0.824]} & \makecell{2.04\\[-1pt]\scriptsize[0.78, 2.86] (36)} & \makecell{6.4\\[-1pt]\scriptsize[3.4, 17.4] (36)} \\ \addlinespace[1.5pt]
\bottomrule
\end{tabular}
\par\medskip
\begin{tabular}{@{}lcccc@{}}
\toprule
Approach (cases) & Edge offset (mm) & Edge level ($\times$ blood) & Terracing (mm) & Roughness (mm) \\
\midrule
\makecell[l]{Taurus-\\base (auto) (38)} & \makecell{+0.42\\[-1pt]\scriptsize[$-$1.72, +1.48] (37)} & \makecell{0.49\\[-1pt]\scriptsize[0.23, 0.91] (22)} & \makecell{0.61\\[-1pt]\scriptsize[0.41, 0.88] (33)} & \makecell{0.61\\[-1pt]\scriptsize[0.47, 1.06] (37)} \\ \addlinespace[1.5pt]
\makecell[l]{Taurus-\\base (38)} & \makecell{+0.80\\[-1pt]\scriptsize[$-$1.72, +1.74] (37)} & \makecell{0.33\\[-1pt]\scriptsize[0.23, 0.51] (22)} & \makecell{0.48\\[-1pt]\scriptsize[0.40, 0.93] (32)} & \makecell{0.56\\[-1pt]\scriptsize[0.46, 0.88] (37)} \\ \addlinespace[1.5pt]
\makecell[l]{Taurus-\\tuned (auto) (38)} & \makecell{+0.60\\[-1pt]\scriptsize[$-$0.20, +1.25] (37)} & \makecell{0.53\\[-1pt]\scriptsize[0.39, 0.58] (24)} & \makecell{0.71\\[-1pt]\scriptsize[0.50, 1.05] (36)} & \makecell{0.50\\[-1pt]\scriptsize[0.44, 0.77] (37)} \\ \addlinespace[1.5pt]
\makecell[l]{Taurus-\\tuned (38)} & \makecell{+0.63\\[-1pt]\scriptsize[$-$0.04, +1.23] (37)} & \makecell{0.52\\[-1pt]\scriptsize[0.38, 0.57] (24)} & \makecell{0.65\\[-1pt]\scriptsize[0.50, 1.07] (36)} & \makecell{0.47\\[-1pt]\scriptsize[0.44, 0.79] (37)} \\ \addlinespace[1.5pt]
\makecell[l]{Taurus-\\hybrid (auto) (38)} & \makecell{+0.54\\[-1pt]\scriptsize[$-$0.09, +1.14] (37)} & \makecell{0.60\\[-1pt]\scriptsize[0.57, 0.62] (25)} & \makecell{0.50\\[-1pt]\scriptsize[0.35, 0.73] (37)} & \makecell{0.43\\[-1pt]\scriptsize[0.35, 0.78] (37)} \\ \addlinespace[1.5pt]
\makecell[l]{Taurus-\\hybrid (38)} & \makecell{+0.76\\[-1pt]\scriptsize[+0.26, +1.46]} & \makecell{0.60\\[-1pt]\scriptsize[0.57, 0.61] (26)} & \makecell{0.38\\[-1pt]\scriptsize[0.23, 0.56]} & \makecell{0.35\\[-1pt]\scriptsize[0.30, 0.55]} \\ \addlinespace[1.5pt]
\makecell[l]{Free-form\\Sonnet (38)} & \makecell{+0.36\\[-1pt]\scriptsize[$-$0.21, +1.27] (36)} & \makecell{0.63\\[-1pt]\scriptsize[0.54, 0.68] (26)} & \makecell{0.31\\[-1pt]\scriptsize[0.20, 0.56] (36)} & \makecell{0.41\\[-1pt]\scriptsize[0.36, 0.55] (36)} \\ \addlinespace[1.5pt]
\makecell[l]{Free-form\\Opus (38)} & \makecell{+1.32\\[-1pt]\scriptsize[+0.23, +2.05] (36)} & \makecell{0.54\\[-1pt]\scriptsize[0.51, 0.57] (25)} & \makecell{0.33\\[-1pt]\scriptsize[0.18, 0.48] (36)} & \makecell{0.38\\[-1pt]\scriptsize[0.29, 0.64] (36)} \\ \addlinespace[1.5pt]
\midrule
Human reference, itself (38) &  & \makecell{0.59\\[-1pt]\scriptsize[0.39, 0.74] (26)} & \makecell{0.47\\[-1pt]\scriptsize[0.31, 0.65]} & \makecell{0.40\\[-1pt]\scriptsize[0.32, 0.52]} \\
\bottomrule
\end{tabular}
\par\medskip
\begin{tabular}{@{}lcccccccc@{}}
\toprule
\makecell[l]{NSD 1 mm\\MSD (mm)} & \makecell{Taurus-\\base (auto)} & \makecell{Taurus-\\base} & \makecell{Taurus-\\tuned (auto)} & \makecell{Taurus-\\tuned} & \makecell{Taurus-\\hybrid (auto)} & \makecell{Taurus-\\hybrid} & \makecell{Free-form\\Sonnet} & \makecell{Free-form\\Opus} \\
\midrule
AortaSeg24 (4) & \makecell{0.589\\[-1pt]11.86} & \makecell{0.590\\[-1pt]11.94} & \makecell{0.712\\[-1pt]1.82} & \makecell{0.713\\[-1pt]1.66} & \makecell{0.712\\[-1pt]1.80} & \makecell{0.716\\[-1pt]1.70} & \makecell{0.273\\[-1pt]7.79} & \makecell{0.590\\[-1pt]3.99} \\ \addlinespace[1pt]
Siriapisith AAA/EVAR (12) & \makecell{0.139\\[-1pt]8.24} & \makecell{0.153\\[-1pt]8.24} & \makecell{0.310\\[-1pt]3.30} & \makecell{0.377\\[-1pt]1.97} & \makecell{0.289\\[-1pt]3.34} & \makecell{0.400\\[-1pt]1.59} & \makecell{0.207\\[-1pt]3.19} & \makecell{0.298\\[-1pt]1.91} \\ \addlinespace[1pt]
MosMedData AAA (8) & \makecell{0.007\\[-1pt]43.65} & \makecell{0.007\\[-1pt]43.65} & \makecell{0.462\\[-1pt]0.90} & \makecell{0.462\\[-1pt]0.90} & \makecell{0.483\\[-1pt]0.75} & \makecell{0.849\\[-1pt]0.65} & \makecell{0.684\\[-1pt]0.76} & \makecell{0.942\\[-1pt]0.38} \\ \addlinespace[1pt]
AVT (4) & \makecell{0.097\\[-1pt]18.08} & \makecell{0.283\\[-1pt]10.43} & \makecell{0.327\\[-1pt]10.50} & \makecell{0.327\\[-1pt]10.50} & \makecell{0.420\\[-1pt]31.08} & \makecell{0.387\\[-1pt]12.25} & \makecell{0.641\\[-1pt]1.59} & \makecell{0.412\\[-1pt]1.31} \\ \addlinespace[1pt]
VMR (4) & \makecell{0.490\\[-1pt]9.88} & \makecell{0.478\\[-1pt]10.04} & \makecell{0.432\\[-1pt]10.55} & \makecell{0.460\\[-1pt]10.56} & \makecell{0.447\\[-1pt]10.57} & \makecell{0.481\\[-1pt]10.16} & \makecell{0.220\\[-1pt]11.93} & \makecell{0.430\\[-1pt]10.92} \\ \addlinespace[1pt]
Jamshidian 4D-CTA (6) & \makecell{0.007\\[-1pt]14.45} & \makecell{0.029\\[-1pt]12.25} & \makecell{0.357\\[-1pt]3.13} & \makecell{0.360\\[-1pt]3.10} & \makecell{0.341\\[-1pt]2.38} & \makecell{0.494\\[-1pt]1.38} & \makecell{0.196\\[-1pt]4.82} & \makecell{0.339\\[-1pt]2.17} \\ \addlinespace[1pt]
\bottomrule
\end{tabular}
\end{table}

\end{table}

\FloatBarrier
\section{Harness failure taxonomy}
\label{app:issues}

Reviewers reported harness defects in their hand-backs, and the score audits, the edit replay and the reviewers of \revB{} and \revC{} added more; the first 65 fall into the groups below, and \cref{tab:issues-v3} lists the ten found with \revC{}. %
The largest groups concern trust in the proposal (mirrored or header-flipped orientation, a trunk on the spine or vertebrae), thresholds that fail in opposite directions on poor and bright contrast, an outer boundary that under-grows the sac or leaks into veins, bowel, psoas and the crura, branch identity and zones, and edit tools whose reach or guard rails did not fit the task.
Others concern latency on whole-body volumes, protocol details (verdicts recorded without opening the latest image, issue codes rejected by the tool), and scoring. %

\begin{table}[h]
  \centering
  \caption{Issues found with \revC{}: 66--68 in the pilot on non-evaluation cases (fixed before the held-out run), 69--75 in the held-out reviews (kept out of the harness design; two generic code faults fixed during the run).
  \src{\texttt{results/harness\_issues.md}}}
  \label{tab:issues-v3}
  \footnotesize
  \begin{tabularx}{\linewidth}{@{}r >{\raggedright\arraybackslash}p{2.1cm} >{\raggedright\arraybackslash}X >{\raggedright\arraybackslash}p{3.1cm}@{}}
    \toprule
    \# & Found in & Problem & Status \\
    \midrule
    66 & pilot, Siriapisith AAA/EVAR and Jamshidian 4D-CTA & blood value contaminated by calcium on a 4D phase; end slices refused by \texttt{interp-outer}; card coordinates on the down-sampled grid; broad outer tools overshoot where no fat plane separates the aorta; false branches in crops & fixed (robust blood, end slices, coordinates) \\
    67 & pilot, whole-body & card coordinates relative to the working box; outer-trimming tools strip the wall ring; false arch-branch specks; working box too narrow; coverage warning on by-design lumen-only iliacs & fixed (wall ring, branch cleanup, box report) \\
    68 & pilot, chest scans & root recipes did nothing; card options computed before the regularization; branch-lumen pieces not removable; edits on the largest volume over 10\,min & fixed (root tracker, options as differences, \texttt{branch-remove}); latency on the largest volume open \\
    69 & held-out, Siriapisith AAA/EVAR and Jamshidian 4D-CTA & missing or collapsed sac runs; growth tools cap giant sacs, key-slice polygons needed; coverage check reported a gap as complete; vertebral branches survive the cleanup; a very bright phase left the lumen empty; guard rails refused calcium hysteresis and the regularization with an endoleak & open \\
    70 & held-out, Siriapisith AAA/EVAR and Jamshidian 4D-CTA & flat-cut sac ends and truncated trunks; growth over 30\,\% refused without forcing; new trunk voxels without a vessel id; tracker recipe crashed on its slice range; giant sac or hematoma undecidable & open; id and crash fixed during the run \\
    71 & held-out, MosMedData AAA & collar needs more than 10\,\% outer removal; side names reversed on header-flipped volumes; lumen on the vertebral body; branch tools refuse clicks inside the trunk; stent graft in a case listed without one & open \\
    72 & held-out, whole-body & tracker fails when the box clips the aorta, when thresholds come from the wrong tissue and on unenhanced scans; unenhanced scan run as enhanced; arch names on far fragments & open \\
    73 & held-out, whole-body & collar on normal-caliber aortas in 4 of 4 cases; wrong vessel names; expected branches not found in 4 of 5; dense calcium missed; command killed at the memory cap & open; cap raised to 16\,GB \\
    74 & held-out, VMR & branches stop early (\texttt{extend\_branch} fixed them); sides reversed on a header-mirrored case; AI outline unreliable & open \\
    75 & held-out, whole-body & collar; superior mesenteric artery truncated or merged into the trunk; limb names; interleaved lumen and thrombus in a false lumen; memory kills on the largest volume; edit applied but not recorded when its render failed & open; record fixed during the run \\
    \bottomrule
  \end{tabularx}
\end{table}

\FloatBarrier
\section{Cost accounting}
\label{app:compute}

All LLM runs were Claude Code sessions under one account, sharing runtime, tools and prompt cache.
For every model response, the transcript records the model id, the timestamps and four token counts: fresh input, cache writes, cache reads and output. %
We price each count at Anthropic's listed token prices for usage as of October 2026 (\cref{tab:prices}), sum the four products for each response and sum the responses of a run to obtain its cost. %
Claude Code writes one-hour cache entries, which are priced at twice the input rate.
A review's cost is its transcript segment plus a share of the session start-up, and it is understated by 3--6\,\% because some output counts are missing, while free-form cost is the client's own report of the same quantity. %
Under a subscription these are cost equivalents, not invoices, and the automated pass, which runs on the CPU, is not priced.
Review time is active transcript time and free-form time the run's wall time. %
We timed the preprocessing that precedes a review separately, on the 38 held-out cases, one case at a time on the otherwise idle workstation (AMD Ryzen 9 3900X, RTX 4090): the vessel proposal on the GPU, then each version's automated pass on the CPU. %
For \taurus{}-hybrid the two took a median of 25\,s per case on Jamshidian 4D-CTA, 26\,s on Siriapisith AAA/EVAR, 76\,s on MosMedData AAA, 195\,s on AortaSeg24, 238\,s on VMR and 310\,s on AVT, of which the proposal took 9--17\,s except on the larger VMR (58\,s) and AVT (143\,s) scans. %
The minutes in \cref{tab:heldout,tab:a5} add these times to the review time, which makes a median of 6.4\,min per case for \revC{} against 16.1\,min for \ffo{}. %
A median Opus review of an AortaSeg24 case read 1.58\,M cached tokens and wrote 1.4\,k output tokens. %

\begin{table}[h]
  \centering
  \caption{Anthropic's listed token prices for usage (October 2026), in US dollars per million tokens.}
  \label{tab:prices}
  \small
  \begin{tabular}{@{}lrrrrr@{}}
    \toprule
    Model & Input & Output & Cache read & Cache write (5 min) & Cache write (1 h) \\
    \midrule
    Claude Opus 5.5 (\texttt{claude-opus-5-5}) & 4.00 & 20.00 & 0.20 & 5.00 & 8.00 \\
    Claude Sonnet 5.5 (\texttt{claude-sonnet-5-5}) & 2.00 & 10.00 & 0.20 & 2.50 & 4.00 \\
    Claude Sonnet 5 (\texttt{claude-sonnet-5}) & 2.00 & 10.00 & 0.20 & 2.50 & 4.00 \\
    \bottomrule
  \end{tabular}
\end{table}

\FloatBarrier
\section{Downstream training on recovered labels}
\label{app:downstream}

To test whether recovered labels can train segmentation models, we trained models on five label sources and evaluated them on held-out human labels of the same dataset. %
The sources are the human labels (L0), \autoA{} (L1), \revA{} (L2), \autoA{} restricted to the cases that the reviewer did not mark \emph{issue} (L2-noedit, which separates the value of the verdict from that of the edits), and \nbu{} (L3).
MosMedData AAA uses 78 training cases (L2-noedit 40) and 20 locked patients, Siriapisith AAA/EVAR uses 130 reviewed contrast-enhanced crops (L2-noedit 56) and the 40 sealed test crops, and AortaSeg24 uses 89 training cases (L2-noedit 27) and 10 locked cases. %
The tissue models use one recipe in the dataset's own classes, 250 epochs and three seeds, and 13 of the 78 MosMedData AAA L1 and L2 training labels are orientation failures. %
The study plan expected L2 $\geq$ L1, L2 within 0.03 of L0, L2-noedit between L1 and L2, and L3 $\geq$ L2. %
\Cref{tab:downstream} gives the results for MosMedData AAA and Siriapisith AAA/EVAR, and the AortaSeg24 runs are still in progress.
Paired tests compare per-case means over the seeds and are unadjusted.

Models trained on \revA{} labels fell well short of models trained on human labels.
On Siriapisith AAA/EVAR, the outer Dice of the L2 models was 0.699 against 0.975 for L0 (paired median $-0.254$; 0 better, 40 worse; $p = 1.8\times10^{-12}$), and on MosMedData AAA their primary metric was 0.522 against 0.804 ($-0.137$; 6/14; $p = 0.006$). %
Review of the training labels helped on Siriapisith AAA/EVAR, where L2 exceeded L1 by a paired median of 0.075 (33/3, $p = 1.2\times10^{-8}$), but not on MosMedData AAA ($-0.000$, 4/6, $p = 0.57$), whose training labels include orientation failures that no edit verb could repair.
Dropping the cases marked \emph{issue} raised the mean on MosMedData AAA from 0.522 to 0.679, although the paired median moved by only $+0.002$ (10/5, $p = 0.19$), while on Siriapisith AAA/EVAR it removed 74 of the 130 training crops and lowered the outer Dice to 0.555 ($-0.109$, 8/32, $p = 5.8\times10^{-7}$).
L3, which starts from the dataset's own labels, matched L0 on Siriapisith AAA/EVAR and reached 0.723 on MosMedData AAA.
Of the plan's four expectations, L3 $\geq$ L2 held on both datasets and L2 $\geq$ L1 on Siriapisith AAA/EVAR only, while the other two held on neither.
These models were trained on \revA{} labels, so whether labels from \revB{} or \revC{}, which scored higher in the held-out comparison, close the gap to human labels remains untested.

\begin{table}[h]
  \centering
  \caption{Models trained on \revA{} labels fell well short of models trained on human labels.
  Mean over the held-out cases of the dataset's primary metric (MosMedData AAA: mean of lumen and calcium Dice over the classes present; Siriapisith AAA/EVAR: outer Dice), mean $\pm$ s.d.\ over three training seeds, per training label source.
  The AortaSeg24 runs are still in progress.
  \textsuperscript{a}Two seeds; the third is still training.}
  \label{tab:downstream}
  \footnotesize
  \setlength{\tabcolsep}{4pt}
  \begin{tabular}{@{}lccccc@{}}
    \toprule
    Dataset (held-out cases) & L0 human & L1 \autoA{} & L2 \revA{} & L2-noedit & L3 reference-seeded \\
    \midrule
    MosMedData AAA (20) & 0.804 $\pm$ 0.039 & 0.507 $\pm$ 0.006 & 0.522 $\pm$ 0.005 & 0.679 $\pm$ 0.025 & 0.723 $\pm$ 0.004 \\
    Siriapisith AAA/EVAR (40) & 0.975 $\pm$ 0.005 & 0.583 $\pm$ 0.008 & 0.699 $\pm$ 0.014 & 0.555 $\pm$ 0.020 & 0.977 $\pm$ 0.000\textsuperscript{a} \\
    \bottomrule
  \end{tabular}
\end{table}

\FloatBarrier
\section{Prompts}
\label{app:prompts}

We summarize the prompts here rather than reproduce them, and the full texts are available from the authors on request.
The harness code and the guides call the edit commands \emph{verbs} and the prepared programs \emph{recipes}, and the appendices keep these names. %

\emph{Free-form task.}
The agent labels one CT volume and works only inside its own directory. %
It reads a label contract that defines the classes and the vessel ids, and it must write three files on the CT's own grid: a tissue map (background, wall, lumen, calcium, thrombus), a vessel map with one id per labeled voxel, and a verdict file with a status (\emph{ok} or \emph{issue}), short issue codes and notes on what it did and what it is unsure about.
The wall is a shell of at most 1.5\,mm that the agent generates outside aneurysmal sections only.
The agent may write and run Python with NumPy, SciPy, scikit-image, nibabel, SimpleITK and Matplotlib and render slices to look at them, but it has no labeling tool and no trained model.
It may not read files outside its directory or ask questions, and it must always leave valid output, because an imperfect label scores better than none.

\emph{Reviewer prompts.}
The three reviewer prompts share one structure. %
The reviewer reads the harness guide, which is its procedure, claims cases one at a time through the harness, opens the review image (and on whole-body cases the vessel map) before any verdict, and edits only through harness commands.
It may read only the guide and the files that harness commands return, so dataset label files and all other folders are off limits.
Each case closes with a verdict, issue codes from a fixed list and short factual notes.
The prompts differ in budget and procedure.
\revA{} has an edit budget of eight set in the job configuration, and \revB{} adds the dossier and the question cards, with at most 20 turns and 12 image opens per case. %
\revC{} allows about 30 turns, 16 image opens and 12 local and 2 global edits, and it adds a second phase of free-form refinement. %
In that phase the reviewer makes the label follow the half-maximum edge, a smooth surface and the complete extent, and checks each edge on zoomed slices before and after every boundary edit.
The \revC{} prompt prefers the regularization and the prepared recipes to the reviewer's own programs, and it asks for reference-disagreement notes on structures that the CT shows but a lumen-only or trunk-only reference may lack.

\fi

\end{document}